\documentclass[runningheads]{llncs}
\usepackage{graphicx}
\usepackage{comment}
\usepackage{amsmath,amssymb} %
\usepackage{color}
\usepackage{url}

\usepackage{caption}
\usepackage{wrapfig}
\usepackage{subfiles}
\usepackage[group-separator={,},output-decimal-marker = {.}, group-minimum-digits={3}]{siunitx}  %
\usepackage{paralist} %
\usepackage{enumitem}
\usepackage{booktabs} %
\usepackage{multirow}
\usepackage{algpseudocode}
\usepackage{algorithm}
\usepackage{tikz}
\usepackage{verbatim}
\usepackage{pgfplots}

\pgfplotsset{compat=newest}
\usetikzlibrary{plotmarks} %
\usetikzlibrary{colorbrewer} %

\usepackage[width=122mm,left=12mm,paperwidth=146mm,height=193mm,top=12mm,paperheight=217mm]{geometry}

\begin{document}
\pagestyle{headings}
\mainmatter

\title{Connecting Vision and Language with\newline Localized Narratives\vspace{-2mm}} %

\titlerunning{Connecting Vision and Language with Localized Narratives}
\author{Jordi Pont-Tuset \and
    Jasper Uijlings\and
    Soravit Changpinyo\and\newline
Radu Soricut \and
Vittorio Ferrari\vspace{-2mm}}
\authorrunning{J. Pont-Tuset et al.}
\institute{Google Research\vspace{-4mm}}

\maketitle
\renewcommand{\floatpagefraction}{.8}%

\newcommand{\eg}[0]{e.g.\ }
\newcommand{\ie}[0]{i.e.\ }

\newcommand{\twocaptions}[5]{
\rule{0mm}{2mm}\\[-2mm]
  \centering\includegraphics[width=#1]{#2}%
  & \centering\arraybackslash \includegraphics[width=#1]{#3}\\
  #4 & #5\rule{0mm}{2mm}\\[-2mm]\\
}

\newif\ifdraft
\draftfalse
\ifdraft
  \newcommand{\Jordi}[1]{{\color{olive}[Jordi] #1}}
  \newcommand{\jasper}[1]{{\color{magenta}[Jasper] #1}}
  \newcommand{\vitto}[1]{{\color{red}[Vitto] #1}}
  \newcommand{\beer}[1]{{\color{cyan}[Beer] #1}}
  \newcommand{\radu}[1]{{\color{orange}[Radu] #1}}
  \newcommand{\att}[1]{{\color{red}#1}}
  \newcommand{\attcr}[1]{{\color{cyan}#1}}
  \newcommand{\new}[1]{#1}
  \newcommand{\newcr}[1]{{\color{blue}#1}}
\else
  \newcommand{\Jordi}[1]{}
  \newcommand{\jasper}[1]{}
  \newcommand{\vitto}[1]{}
  \newcommand{\beer}[1]{}
  \newcommand{\radu}[1]{}
  \newcommand{\att}[1]{#1}
  \newcommand{\attcr}[1]{#1}
  \newcommand{\new}[1]{#1}
  \newcommand{\newcr}[1]{#1}
\fi

\newcommand{\mypartop}[1]{\vspace{0mm}\noindent\textbf{#1}}
\newcommand{\mypar}[1]{\vspace{0.5em}\noindent\textbf{#1}}

\newcommand{\cls}[1]{\texttt{#1}}  %
\newcommand{\coco}{COCO}
\newcommand{\cococite}{\cite{lin14eccv,chen15arxiv}}
\newcommand{\oid}{Open~Images}
\newcommand{\flickr}{Flickr30k}
\newcommand{\ade}{ADE20K}
\newcommand{\vg}{Visual Genome}
\newcommand{\oidcite}{\cite{kuznetsova18arxiv,openimages}}

\newcommand{\numimages}{\attcr{\num{848749}}}
\newcommand{\numimagesshort}{\attcr{\num{849}k}}

\newcommand{\numimagescoco}{\attcr{\num{123287}}}
\newcommand{\numimagescocoshort}{\attcr{\num{123}k}}

\newcommand{\numimagesoidtrain}{\attcr{\num{504413}}}
\newcommand{\numimagesoidtrainshort}{\attcr{\num{504}k}}

\newcommand{\numimagesoidvaltest}{\attcr{\num{167056}}}
\newcommand{\numimagesoidvaltestshort}{\attcr{\num{167}k}}

\newcommand{\numimagesoid}{\attcr{\num{671469}}}
\newcommand{\numimagesoidshort}{\attcr{\num{671}k}}

\newcommand{\numimagesade}{\attcr{\num{20210}}}
\newcommand{\numimagesadeshort}{\attcr{\num{20}k}}

\newcommand{\numimagesflickr}{\attcr{\num{31783}}}
\newcommand{\numimagesflickrshort}{\attcr{\num{32}k}}

\newcommand{\numcaptions}{\attcr{\num{873107}}}

\newcommand{\captionlength}{\attcr{\num{36.5}}}
\newcommand{\captionlengthcoco}{\attcr{\num{41.8}}}
\newcommand{\captionlengthoidtrain}{\attcr{\num{35.5}}}

\newcommand{\annotationtime}{\attcr{\num{144.7}}}
\newcommand{\transcriptiontime}{\attcr{\num{104.3}}}
\newcommand{\transcriptiontimepercent}{\attcr{\num{72}}}
\newcommand{\narrationtime}{\attcr{\num{40.4}}}
\newcommand{\narrationtimepercent}{\attcr{\num{28}}}

\begin{abstract}
We propose Localized Narratives, a new form of multimodal
image annotations connecting vision and language.
We ask annotators to describe an image with their voice while simultaneously
hovering their mouse over the region they are describing.
Since the voice and the mouse pointer are synchronized, we can localize every
single word in the description.
This dense visual grounding takes the form of a mouse trace segment per word and
is unique to our data.
We annotated \numimagesshort{} images with Localized Narratives: \newcr{the whole
\coco{}, \flickr{}, and \ade{} datasets}, and \numimagesoidshort{} images of \oid{},
\newcr{all of} which we make publicly available.
We provide an extensive analysis of these annotations showing they are
diverse, accurate, and efficient to produce.
We also demonstrate their utility on the application of controlled image captioning.
\vspace{-2mm}
\end{abstract}

\section{Introduction}\label{sec:introduction}
\vspace{-1mm}
Much of our language is rooted in the visual world around us.
A popular way to study this connection is through Image Captioning, which uses
datasets where images are paired with human-authored textual
captions~\cite{chen15arxiv,young14tacl,sharma18acl}.
Yet, many researchers want deeper visual grounding which links specific words
in the caption to specific regions in the
image~\cite{liu17aaai,lu18cvpr,rohrbach18emnlp,selvaraju19iccv}.
Hence \flickr{} Entities~\cite{plummer17ijcv} enhanced \flickr{}~\cite{young14tacl}
by connecting the nouns from the captions to bounding boxes in the images.
But these connections are still sparse and important aspects remain ungrounded,
such as words capturing relations between nouns (as ``holding'' in ``a woman
holding a balloon'').
\vg{}~\cite{krishna17ijcv} provides short descriptions of regions%
, thus words are not individually grounded either.

\begin{figure}[t]
  \vspace{-2mm}
  \centering
  \begin{minipage}{0.7\linewidth}
      \includegraphics[width=\linewidth]{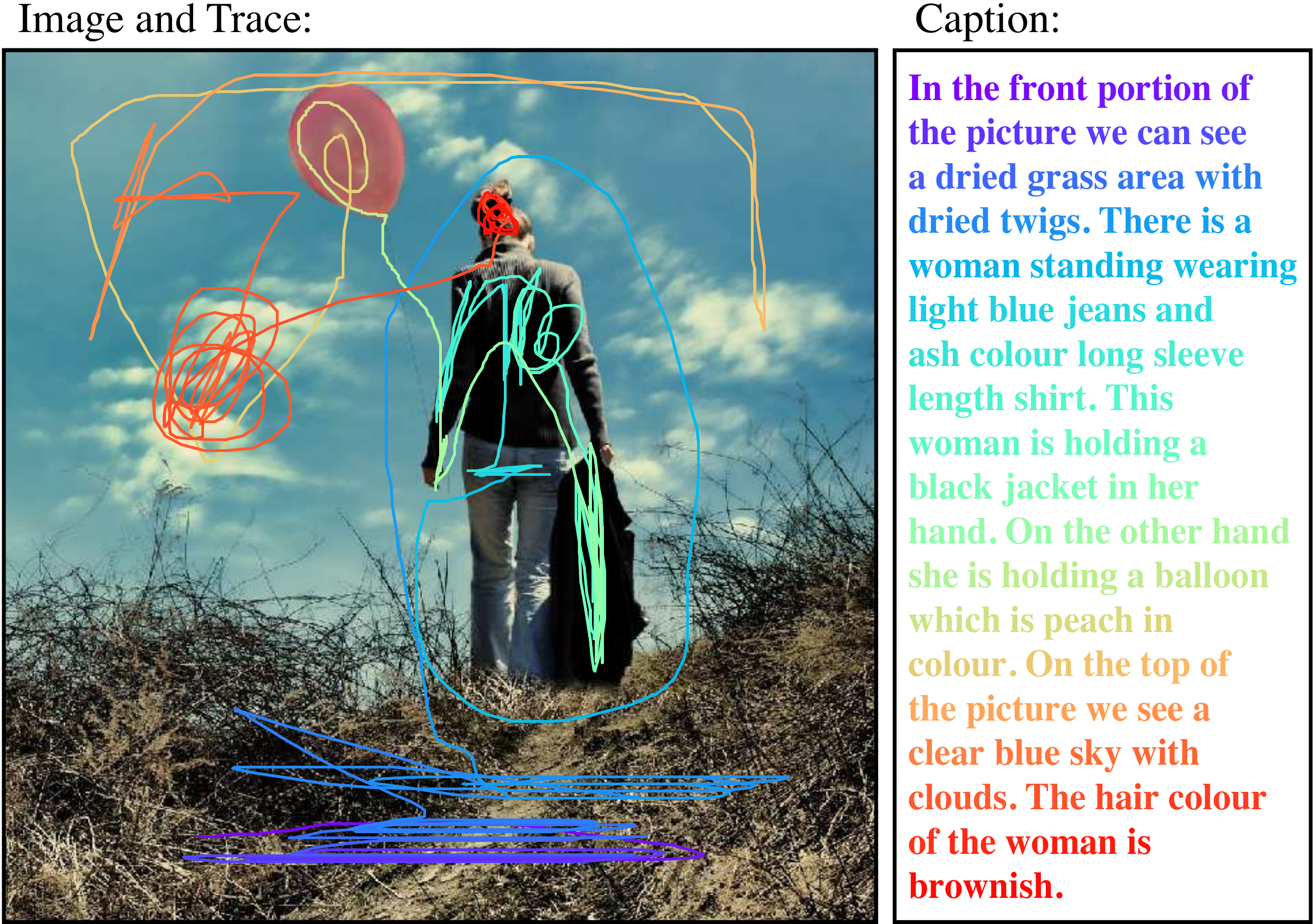}
  \end{minipage}
  \begin{minipage}{0.29\linewidth}
  \includegraphics[width=\linewidth]{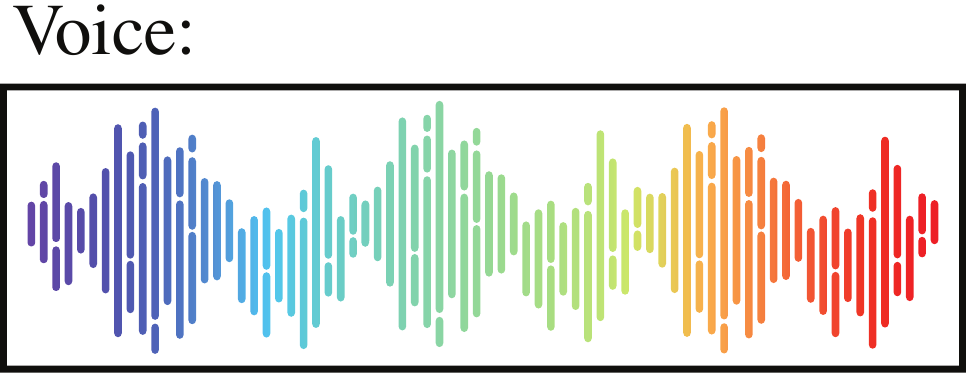}
  \vspace{4.15mm}
  \caption{\small\textbf{Localized Narrative example}: Caption, voice, and mouse trace synchronization represented by a color gradient \protect\includegraphics[width=8mm]{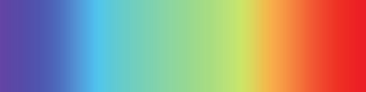}.
  \newcr{The project website~\cite{locnarr-website} contains a visualizer with many live examples.}}
  \label{fig:loc_narr_intro_example}
  \end{minipage}
  \vspace{-7mm}
\end{figure}

\new{In this paper we propose Localized Narratives,
a new form of multimodal image annotations in which
we ask annotators to describe an image with their voice while simultaneously
hovering their mouse over the region they are describing.
Figure~\ref{fig:loc_narr_intro_example} illustrates the process:
the annotator says ``woman'' while using their mouse to indicate her spatial
extent, thus providing visual grounding for this noun.
Later they move the mouse from the woman to the balloon following its string,
saying ``holding''.
This provides direct visual grounding of this relation.
They also describe attributes like ``clear blue sky'' and ``light blue jeans''.
Since voice is \emph{synchronized} to the mouse pointer, we can determine
the image location of every single word in the description.
This provides dense visual grounding in the form of a mouse trace segment for
each word, which is unique to our data.}

In order to obtain written-word grounding, we additionally need to transcribe
the voice stream.
We observe that automatic speech recognition~\cite{graves13icassp,amodei16icml,ravanelli19icassp}
typically results in imperfect transcriptions.
To get data of the highest quality, we ask annotators to transcribe their
own speech, immediately after describing the image.
This delivers an accurate transcription, but without temporal synchronization
between the mouse trace and the written words.
To address this issue, we perform a sequence-to-sequence alignment between
automatic and manual transcriptions, which leads to accurate {\em and} temporally
synchronized captions.
Overall, our annotation process tightly connects four modalities:
the image, its spoken description, its textual description, and the mouse trace.
Together, they provide dense grounding between language and vision.

Localized Narratives is an efficient annotation protocol.
Speaking and pointing to describe things comes naturally to humans~\cite{kahneman73attention,oviatt03book}.
Hence this step takes little time (\narrationtime{}~sec.\ on average).
The manual transcription step takes \transcriptiontime{}~sec., for a total of \annotationtime{}~sec.
This is lower than the cost of related grounded captioning datasets~\flickr{}~Entities
and \vg{}~\cite{krishna17ijcv,plummer17ijcv},
which were made by more complicated annotation processes and involved manually
drawing bounding boxes (Sec.~\ref{subsec:dataset-collection} -- Annotation Cost).
Moreover, if automatic speech recognition improves in the future it might be
possible to skip the manual transcription step, making our approach even more efficient.

We collected Localized Narratives at scale:
\newcr{
we annotated the whole~\coco{}~\cite{lin14eccv} (\numimagescocoshort{} images), %
\ade{}~\cite{zhou19ijcv} (\numimagesadeshort{})
and \flickr{}~\cite{young14tacl} (\numimagesflickrshort{}) datasets,
as well as \numimagesoidshort{} images of \oid{}~\cite{kuznetsova18arxiv}.
We make the Localized Narratives for these \att{\numimages{}} images publicly available~\cite{locnarr-website}.
}
We provide an extensive analysis (Sec.~\ref{sec:analysis}) and show that:
(i) Our data is rich: we ground all types of words (nouns, verbs, prepositions, etc.\@),
and our captions are \newcr{substantially} longer than in \newcr{most} %
previous datasets~\cite{chen15arxiv,young14tacl,krishna17ijcv,sharma18acl}.
(ii) Our annotations are diverse both in the language modality
(\eg{}caption length varies widely with the content of the image) and in the
visual domain
(different pointing styles and ways of grounding relationships).
(iii) Our data is of high quality: the mouse traces match well the location of the objects,
the words in the captions are semantically correct, and the manual transcription
is accurate.
(iv) Our annotation protocol is more efficient than for related grounded
captioning datasets~\cite{krishna17ijcv,plummer17ijcv}.

Since Localized Narratives provides four synchronized modalities, it enables
many applications (Tab.~\ref{tab:tasks}).
We envision that having each word in the captions grounded, beyond the sparse
set of boxes of previous datasets~\cite{plummer17ijcv,kazemzadeh14emnlp,mao16cvpr,krishna17ijcv},
will enable richer results in many of these tasks and open new doors for tasks
and research directions that would not be possible with previously existing annotations.
As a first example, we show how to use the mouse trace as a fine-grained
control signal for a user to request a caption on a particular image (Sec.~\ref{sec:controlled-image-captioning}).
Mouse traces are a more natural way for humans to provide a sequence of
grounding locations, compared to drawing a list of bounding boxes~\cite{cornia19cvpr}.
We therefore envision its use as assistive technology for people with \newcr{imperfect vision}.
In future work, the mouse trace in our Localized Narratives could be used
as additional attention supervision at training time, replacing or
\begin{table}[t!]
  \centering
  \resizebox{\linewidth}{!}{%
  \begin{tabular}{@{}c@{}c@{}c@{\hspace{2mm}}c@{\hspace{10mm}}p{97mm}}
    \toprule
    Image & Text & Speech & Grounding & Task\\
    \midrule
    \color{blue}In & \color{red}Out & - & - & Image captioning~\cite{vinyals15cvpr,xu15icml,vinyals16pami},
    \newcr{Paragraph generation~\cite{krause17cvpr,yan20arxiv,ziegler19arxiv}}\\
    \color{red}Out & \color{blue}In & - & - & Text-to-image Generation~\cite{reed16nips,tan19cvpr,yin19cvpr_b}\\
    \color{blue}In & \color{red}Out & - & \color{red}Out & Dense image captioning~\cite{johnson16cvpr,yin19cvpr,kim19cvpr}, Dense relational captioning~\cite{kim19cvpr}\\
    \color{blue}In & \color{red}Out & - & \color{blue}In & Controllable and Grounded Captioning~\cite{cornia19cvpr}\\
    \color{blue}In & \color{blue}In & - & \color{red}Out & Phrase grounding~\cite{dogan19cvpr}\\
    \color{blue}In & {\color{blue}In} + \color{red}Out & - & - & Visual Question Answering~\cite{antol15iccv,malinowski15iccv,hudson19cvpr}\\
    \color{blue}In & {\color{blue}In} + \color{red}Out & - & \color{red}Out & Referring Expression Recognition~\cite{kazemzadeh14emnlp,mao16cvpr,cirik18naacl}\\
    \color{blue}In & - & \color{blue}In & \color{red}Out & Discover visual objects and spoken words from raw sensory input~\cite{harwath18eccv}\\
    - & \color{blue}In & \color{red}Out & - & Speech recognition~\cite{graves13icassp,amodei16icml,ravanelli19icassp}\\
    - & \color{red}Out & \color{blue}In & - & Speech synthesis~\cite{oord2016arxiv,kalchbrenner18icml,mehri16iclr}\\
    \color{red}Out & \color{blue}In & - & \color{blue}In & Image generation/retrieval from traces\\
    \color{blue}In & \color{red}Out & \color{blue}In & \color{blue}In & Grounded speech recognition\\
    \color{blue}In & - & \color{blue}In & \color{red}Out & Voice-driven environment navigation\\
    \bottomrule
  \end{tabular}}\\[3mm]
  \caption{\small\textbf{Tasks enabled by Localized Narratives}.
  Each row represents different uses of the four elements in a Localized Narrative: image, textual caption, speech, and grounding (mouse trace); labeled as being input ({\color{blue}In}) or output ({\color{red}Out}) for each task.}
  \vspace{-10mm}
  \label{tab:tasks}
\end{table}
complementing the self-supervised attention mechanisms typical of modern
systems~\cite{anderson18cvpr,sharma18acl,changpinyo2019decoupled,yu2019multimodal,xu15icml}.
This might train better systems and improve captioning
performance at test time, when only the image is given as input.
\newcr{Alternatively, our mouse traces could be used at test time only, to inspect whether current spatial attention models activate on the same image regions that humans associate with each word.}

Besides image captioning, Localized Narratives are a natural fit for:
(i) image generation: the user can describe which image they want to generate by talking and moving their mouse to indicate the position of objects (demonstration in App.~\ref{appendix:image-generation-using-traces});
(ii) image retrieval: the user naturally describes the content of an image they are looking for, in terms of both what and where;
(iii) grounded speech recognition: \new{considering} the content of an image would allow better speech transcription, \eg{}`plant' and `planet' are easier to distinguish in the visual than in the voice domain;
(iv) voice-driven environment navigation: the user describes where they want to navigate to, using relative spatial language.

To summarize, our paper makes the following contributions:
(i) We introduce Localized Narratives, a new form of multimodal image annotations
where every word is localized in the image with a mouse trace segment;
(ii) We use Localized Narratives to annotate \numimages{}
images and provide a thorough analysis of the data.
(iii) We demonstrate the utility of our data for controlled image captioning. %

\section{Related Work}\label{sec:related}
\begin{figure}[t!]
  \vspace{-3mm}
  \includegraphics[width=\linewidth]{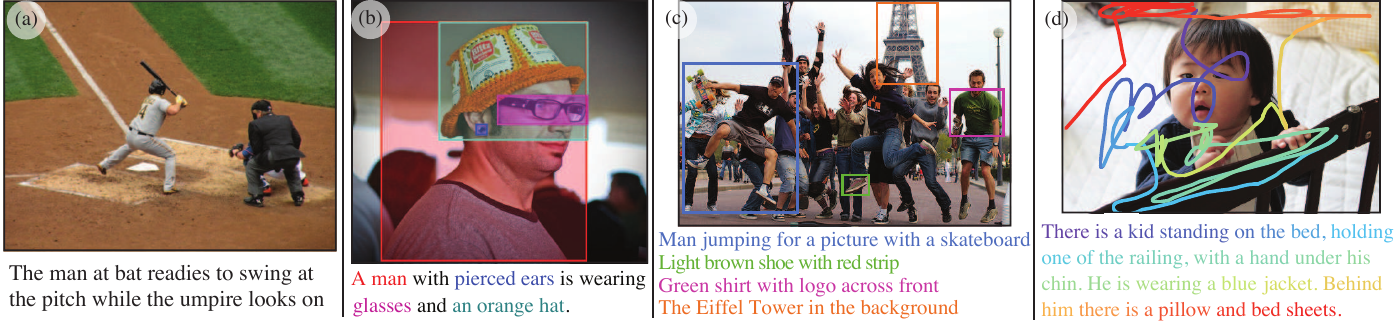}\\[-15pt]
  \caption{\small\textbf{Sample annotations} from (a) \coco{} Captions~\cite{chen15arxiv},
  (b) \flickr{}~ Entities~\cite{plummer17ijcv}, (c) \vg{}~\cite{krishna17ijcv},
  and (d) Localized Narratives (Ours).
  For clarity, (b) shows a subset of region descriptions and (d) shows a shorter-than-average Localized Narrative.}
  \vspace{-3mm}
  \label{fig:sample_datasets}
\end{figure}

\vspace{-3mm}
\mypar{Captioning Datasets.}
\newcr{Various} annotation efforts connect vision and language via captioning (Tab.~\ref{tab:datasets}).
We focus on how their captions are grounded, as this is the key
differentiating factor of Localized Narratives from these works.
As a starting point, classical image captioning~\cite{chen15arxiv,young14tacl,sharma18acl}
\newcr{and visual paragraph generation~\cite{krause17cvpr}}
simply provide a whole caption for the whole image~(Fig.~\ref{fig:sample_datasets}(a)).
This lack of proper grounding was shown to be problematic~\cite{liu17aaai,lu18cvpr,rohrbach18emnlp,selvaraju19iccv}.

\begin{table}[t!]
  \centering
  \resizebox{\linewidth}{!}{%
  \begin{tabular}{l@{\hspace{5mm}}c@{\hspace{2mm}}r@{\hspace{2mm}}r@{\hspace{2mm}}r@{\hspace{2mm}}r}
    \toprule
    Dataset & Grounding & Num. captions & Num. images   & Words/capt. & \\
    \midrule
    \coco{} Captions~\cite{chen15arxiv}  & Whole capt. $\rightarrow$ Whole im. & \num{616767} & \num{123287} & \num{10.5} \\
    Conceptual Capt.~\cite{sharma18acl}  & Whole capt. $\rightarrow$ Whole im. & \num{3334173} & \num{3334173} & \num{10.3} \\
    \newcr{Stanford Vis. Par.~\cite{krause17cvpr}}  & \newcr{Whole capt. $\rightarrow$ Whole im.} & \newcr{\num{19561}} & \newcr{\num{19561}} & \newcr{\num{67.5}}  \\
    \midrule
    ReferIt~\cite{kazemzadeh14emnlp}     & Short phrase $\longrightarrow$ Region & \num{130525} & \num{19894} & \num{3.6} \\
    Google Refexp~\cite{mao16cvpr}       & Short phrase $\longrightarrow$ Region & \num{104560} & \num{26711} & \num{8.4} \\
    \vg{}~\cite{krishna17ijcv}           & Short phrase $\longrightarrow$ Region & \num{5408689} & \num{108077} & \num{5.1} \\
    \midrule
    \flickr{}~ Ent.~\cite{plummer17ijcv} & Nouns $\longrightarrow$ Region & \num{158915} & \num{31783} & \num{12.4} \\
    Loc. Narr. (Ours)                    & Each word $\longrightarrow$ Region & \numcaptions{} & \numimages{} & \captionlength{} \\
    \bottomrule
  \end{tabular}}\\[2mm]
  \caption{\small\textbf{Datasets connecting vision and language via image captioning},
  compared with respect to their type of grounding, scale, and caption length.
  \newcr{Num. captions is typically higher than num. images because of replication (\ie{}several annotators writing a caption for the same image).}
  }
  \label{tab:datasets}
\end{table}

\flickr{} Entities~\cite{plummer17ijcv} annotated the nouns mentioned in the captions of \flickr{}~\cite{young14tacl} and drew their
bounding box in the image (Fig.~\ref{fig:sample_datasets}(b)): the grounding
is therefore from nouns to regions (including their attached adjectives, Tab.~\ref{tab:datasets}).
\vg{}~\cite{krishna17ijcv} and related previous efforts~\cite{kazemzadeh14emnlp,mao16cvpr}
provide short phrases describing regions in the images (Fig.~\ref{fig:sample_datasets}(c)): grounding is therefore at the phrase level (Tab.~\ref{tab:datasets}).
While Visual Genome uses these regions as a seed to generate a scene graph, where each
node is grounded in the image, the connection between the region descriptions and the scene graph is not explicit.

In Localized Narratives, in contrast,
{\em every word} is grounded to a specific region in the image represented by
its trace segment (Fig.~\ref{fig:sample_datasets}(d)).
This includes all types of words (nouns, verbs, adjectives, prepositions, etc.\@),
in particular valuable spatial-relation markers (``above'', ``behind'', etc.\@)
and relationship indicators (``riding'', ``holding'', etc.\@).
Another disadvantage of \flickr{} Entities and \vg{} is that their annotation processes
require manually drawing \new{many}
bounding boxes a posteriori, 
\new{which is unnatural and time-consuming}
compared to our simpler and more natural protocol
(Sec.~\ref{subsec:dataset-collection} -- Annotation Cost).

SNAG~\cite{vaidyanathan18acl} is a proof of concept where annotators describe
images using their voice while their gaze is tracked using specialized hardware.
This enables inferring the image location they are looking at.
As a consequence of the expensive and complicated setup, only \num{100} images
were annotated.
In our proposed Localized Narratives, instead, we collect the data using just
a mouse, a keyboard, and a microphone as input devices, which are commonly available.
This  allows us to annotate a much larger set of images
(\att{\numimages{}} to date).

In the video domain, ActivityNet-Entities~\cite{zhou19cvpr} adds visual grounding
to the ActivityNet Captions, also in two stages 
where boxes were drawn a posteriori.

\mypar{Annotation using Voice.}
\new{A few recent papers use voice as an input modality for}
computer vision tasks~\cite{dai16thesis,vasudevan17cvpr,vaidyanathan18acl,harwath18eccv,gygli19cvpr,gygli19ijcv}.
The closest work to ours~\cite{gygli19ijcv} uses voice to simultaneously annotate
the class name and the bounding box of an object instance in an image.
With Localized Narratives we bring it to the next level by producing richer
annotations both in the language and vision domains with long free-form captions
associated to synchronized mouse traces.

\newcr{In the video domain, EPIC-KITCHENS~\cite{damen20pami} contains videos of daily kitchen 
activities collected with a head-mounted camera.
The actions were annotated with voice, manually transcribed, and time-aligned using 
YouTube's automatic closed caption alignment tool.}

\vspace{-3mm}
\section{Annotation Process}
\label{sec:generation}
\vspace{-3mm}
The core idea behind the Localized Narratives annotation protocol is to ask the
annotators to describe the contents of the image using their voice while hovering
their mouse over the region being described.
Both voice and mouse location signals are timestamped, so we know where the annotators
are pointing while they are speaking every word.

\begin{figure}[t!]
  \vspace{-3mm}
  \centering
  \includegraphics[width=\linewidth]{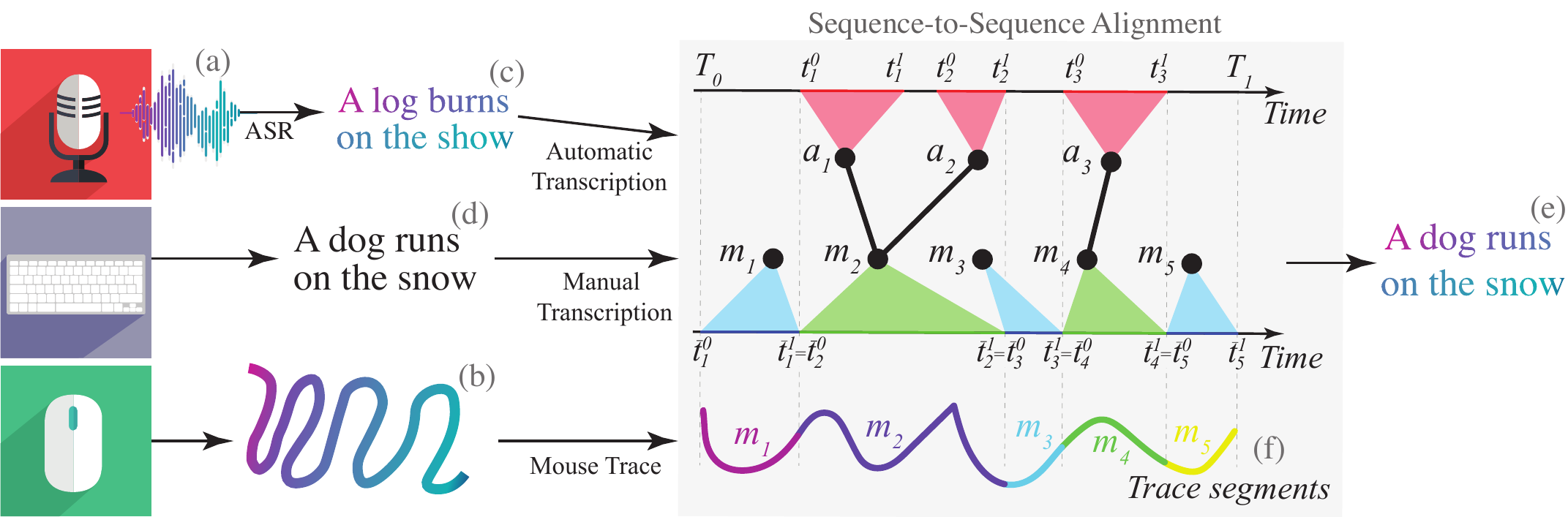}
  \caption{\small\textbf{Localized Narratives annotation}:
  We align the automatic transcription (c) to the manual one (d) to transfer the
  timestamps from the former to the latter, resulting in a transcription that is both
  accurate and timestamped (e).
  To do so, we perform a sequence-to-sequence alignment (gray box) between
  $a_i$ and $m_j$ (black thick lines).
  The timestamps of matched words $m_j$ are defined as the segment (green)
  containing the original timestamps (red) of the matched words $a_i$.
  Unmatched words $m_j$ get assigned the time segments in between matched
  neighboring words (blue).
  These timestamps are transferred to the mouse trace and define the trace segment
  for each word $m_j$.}
  \vspace{-3mm}
  \label{fig:alignment}
\end{figure}

Figure~\ref{fig:alignment} shows voice (a) and mouse trace data (b), where the color
gradient represents temporal synchronization.
We summarize how to process this data to produce a Localized Narrative example.
First, we apply an Automatic Speech Recognition (ASR) algorithm and get a synchronized,
but typically imperfect, transcription (c).
After finishing a narration, the annotators transcribe their own recording, which
gives us an accurate caption, but without synchronization with the mouse trace
(d).
Finally, we obtain a correct transcription with timestamps by performing
sequence-to-sequence alignment between the manual and automatic transcriptions
(e).
This time-stamped transcription directly reveals which trace segment corresponds
to each word in the caption (f), and completes the creation of a Localized Narrative instance.
Below we describe each step in detail.

\mypar{Annotation Instructions.}
One of the advantages of Localized Narratives is that it is a natural task
for humans to do: speaking and pointing at what we are describing is a common
daily-life experience~\cite{kahneman73attention,oviatt03book}.
This makes it easy for annotators to understand the task and perform as
expected, while increasing the pool of qualified annotators for the task.
The instructions we provide are concise:
\\[-2mm]

\setlength{\fboxsep}{0pt}
\vspace{-2mm}\noindent\fbox{
\begin{minipage}{\linewidth}
  \centering
  \begin{minipage}{\linewidth}
    \centering\footnotesize
    \vspace{2mm}
    \textbf{Use the mouse to point at the objects in the scene.\\
  Simultaneously, use your voice to describe what you are pointing at.}
    \vspace{2mm}
  \end{minipage}
  \begin{minipage}{0.95\linewidth}
    \footnotesize
    \begin{asparaitem}
      \item Focus on concrete objects (\eg{}cow, grass, person, kite, road, sky).
      \item Do not comment on things you cannot directly see in the image (\eg{}feelings
    that the image evokes, or what might happen in the future).
      \item Indicate an object by moving your mouse over the whole object, roughly
    specifying its location and size.
      \item Say the relationship between two objects while you move the mouse between
    them, \eg{}``a man \textit{is flying} a kite'', ``a bottle \textit{is on} the table''.
      \item If relevant, also mention attributes of the objects (\eg{}\textit{old} car).
    \end{asparaitem}
  \end{minipage}
  \vspace{2mm}
\end{minipage}}
\vspace{-1mm}

\mypar{Automatic and Manual Transcriptions.}
We apply an ASR algorithm~\cite{googlecloudapi} to obtain an automatic
transcription of the spoken caption, which is timestamped but typically contains
transcription errors.
To fix these errors, we ask the annotators to manually transcribe their own recorded
narration.
Right after they described an image, the annotation tool plays their own voice
recording accompanied by the following instructions:
\\[-2mm]

\setlength{\fboxsep}{0pt}
\noindent\fbox{
\begin{minipage}{\linewidth}
  \centering
  \begin{minipage}{\linewidth}
    \centering\footnotesize
    \vspace{2mm}
    \textbf{Type literally what you just said.}
    \vspace{2mm}
  \end{minipage}
  \begin{minipage}{0.95\linewidth}
    \footnotesize
    \begin{asparaitem}
      \item Include filler words if you said them (\eg{}``I think'', ``alright'') but not filler
      sounds (\eg{}``um'', ``uh'', ``er'').
      \item Feel free to separate the text in multiple sentences and add punctuation.
    \end{asparaitem}
  \end{minipage}
  \vspace{2mm}
\end{minipage}}
\vspace{-1mm}

The manual transcription is accurate but not timestamped, so we cannot
associate it with the mouse trace to recover the grounding of each word.

\mypar{Transcription Alignment.}
We obtain a correct transcription with timestamps by performing a
sequence-to-sequence alignment between the manual and automatic transcriptions
(Fig.~\ref{fig:alignment}).

Let $\mathbf{a}\!=\!\{(a_1,\dots,a_{|\mathbf{a}|}\}$ and
$\mathbf{m}\!=\!\{m_1,\dots,m_{|\mathbf{m}|}\}$ be the automatic and manual
transcriptions of the spoken caption, where $a_i$ and $m_j$ are individual words.
$a_i$ is timestamped: let $[t_i^0, t_i^1]$ be the time segment during which
$a_i$ was spoken.
Our goal is to  align $\mathbf{a}$ and $\mathbf{m}$ to transfer the timestamps
from the automatically transcribed words $a_i$ to the manually provided $m_j$.

To do so, we apply Dynamic Time Warping~\cite{kruskal99} between $\mathbf{a}$ and $\mathbf{m}$.
Intuitively, we look for a matching function $\mu$ that assigns each word $a_i$ to a
word $m_{\mu(i)}$, such that if $i_2\!>\!i_1$ then $\mu(i_2)\!\geq\!\mu(i_1)$
(it preserves the order of the words).
Note that $\mu$ assigns each $a_i$ to exactly one $m_j$,
  but $m_j$ can match to zero or multiple words in $\mathbf{a}$.
We then look for the optimal matching $\mu^*$ such that:
\vspace{-2mm}%
{\small
\begin{equation}
  \mu^*=\arg\min_\mu D_\mu(\mathbf{a},\mathbf{m}) \qquad\qquad
  D_\mu(\mathbf{a},\mathbf{m})=\sum_{i=1}^{|\mathbf{a}|}d(a_i, m_{\mu(i)})
  \label{eq:distance}
\vspace{-2mm}
\end{equation}}%
where $d$ is the edit distance between two words, \ie{}the number of
character inserts, deletes, and replacements required to get
from one word to the other. $D_{\mu^*}(\mathbf{a},\mathbf{m})$ provides the optimal matching score (used below to assess quality).\\[-0.8em]

Given $\mu^*$, let the set of matches for $m_j$ be defined as $A_j=\left\{i\,|\,\mu^*(i)\!=\!j\right\}$.
The timestamp $[\bar{t}_j^0, \bar{t}_j^1]$ of word $m_j$ in the manual transcription
is the interval spanned by its matching words (if any) or to the time between
neighboring matching words (if none).
\Jordi{Candidate to remove}%
Formally:
{\small%
\begin{equation}
\bar{t}_j^0=\left\{\begin{array}{l}
\min\left\{t_i^0\,|\,i\in A_j\right\}\text{ if }A_j\neq\emptyset\text{,}\\
\max\left\{t_i^1\,|\,i\in A_k\,|\,k<j\right\}\text{ if }\exists\,k<j \text{ s.t. }A_k\neq\emptyset\\
T^0 \text{ otherwise,}
\end{array} \right.
\end{equation}}%
{\small%
\[
\bar{t}_j^1=\left\{\begin{array}{l}
                 \max\left\{t_i^1\,|\,i\in A_j\right\}\text{ if }A_j\neq\emptyset\text{,}\\
                 \min\left\{t_i^0\,|\,i\in A_k\,|\,k>j\right\}\text{ if }\exists\,k>j \text{ s.t. }A_k\neq\emptyset\\
                 T^1 \text{ otherwise,}
\end{array} \right.
\]}%
where $T^0$ is the first time the mouse pointer \textit{enters} the image and $T^1$ is
the last time it \textit{leaves} it.
Finally, we define the \textit{trace segment} associated with a word $m_j$ as
the segment of the mouse trace spanned by the time interval $[\bar{t}_j^0,
\bar{t}_j^1]$ (Fig.~\ref{fig:alignment}).

\mypar{Automatic quality control.}
To ensure high-quality annotations, 
we devise an automatic quality control mechanism by leveraging the fact
that we have a double source of voice transcriptions:
the manual one given by the annotators ($\mathbf{m}$) and
the automatic one given by the ASR system ($\mathbf{a}$, Fig.~\ref{fig:alignment}).
We take their distance $D_{\mu^*}(\mathbf{a},\mathbf{m})$ in the optimal
alignment $\mu^*$ as a quality control metric (Eq.~\eqref{eq:distance}).
A high value of $D_{\mu^*}$ indicates large discrepancy between the two transcriptions,
which could be caused by the annotator having wrongly transcribed the text,
or due to the ASR failing to recognize the annotators' spoken words.
In contrast, a low value of $D_{\mu^*}$ indicates that the transcription is
corroborated by two sources.
In practice, we manually analyzed a large number of annotations at different values
of $D_{\mu^*}$ and choose a specific threshold below which essentially all
transcriptions were correct. We discarded all annotations above this threshold.

In addition to this automatic quality control, we also evaluate the quality of the
annotations in terms of semantic accuracy, visual grounding accuracy, and quality of manual voice transcription
(Sec.~\ref{subsec:dataset-quality}).

\section{Dataset Collection, Quality, and Statistics}
\label{sec:analysis}

\subsection{Dataset collection}\label{subsec:dataset-collection}

\mypartop{Image Sources and Scale.}
\newcr{We annotated a total of \numimages{} images with Localized Narratives over 4 datasets:
(i) \coco{}~\cococite{} (train and validation, \numimagescocoshort{} images);
(ii) \flickr{}~\cite{young14tacl} (train, validation, and test, \numimagesflickrshort{});
(iii) \ade{}~\cite{zhou19ijcv} (train and validation, \numimagesadeshort{});
(iv)  \oid{} (full validation and test, \numimagesoidvaltestshort{}, and part of train, \numimagesoidtrainshort{}).
}
\newcr{For \oid{}}, to enable cross-modal applications, we selected images for which object
segmentations~\cite{benenson19cvpr}, bounding boxes or visual relationships~\cite{kuznetsova18arxiv}
are already available.
We annotated \attcr{\num{5000}} \newcr{randomly selected} \coco{} images with
replication~5 (\ie{}5 different annotators annotated each image).
Beyond this, we prioritized having a larger set
covered, so the rest of images were annotated with replication~1.
All analyses in the remainder of this section are done on the \newcr{full set of \numimagesshort{} images,
unless otherwise specified}.

\mypar{Annotation Cost.}
Annotating one image with Localized Narratives takes \annotationtime{} seconds
on average.
\new{We consider this a relatively low cost given the amount of information harvested,
and it allows data collection at scale.
}
Manual transcription takes up the majority of the time (\transcriptiontime{} sec., \transcriptiontimepercent{}\%),
while the narration step only takes \narrationtime{} seconds (\narrationtimepercent{}\%).
In the future, when ASR systems improve further, manual transcription could be
skipped and Localized Narratives could become even faster thanks to our core idea
of using speech.

\new{
To put our timings into perspective, we can roughly compare to \flickr{} Entities~\cite{plummer17ijcv},
which is the only work we are aware of that reports annotation times.
They first manually identified which words constitute entities, which took
235 seconds per image.
In a second stage, annotators drew bounding boxes for these selected entities,
taking 408 seconds (8.7 entities per image on average).
This yields a total of 643 seconds per image, without counting the time to write
the actual captions (not reported).
This is \attcr{$4.4 \times$} slower than the total annotation cost of our method, which includes
the grounding of \attcr{\num{10.8}} nouns
per image and the writing of the caption.
The \vg{}~\cite{krishna17ijcv} dataset was also annotated by a complex multi-stage
pipeline, also involving drawing a bounding box for each phrase describing a
region in the image.
}

\subsection{Dataset Quality}\label{subsec:dataset-quality}
\new{To ensure high quality, Localized Narratives was made by \attcr{\num{156}} professional annotators
working full time on this project.
Annotator managers did frequent manual inspections to keep quality consistently high.
In addition, we used an automatic quality control mechanism to ensure that the
spoken and written transcriptions match (Sec.~\ref{sec:generation} -- Automatic quality control).
In practice, we placed a high quality bar, which resulted in discarding \attcr{23.5\%}
of all annotations.
Below we analyze the quality of the annotations that remained after this automatic discarding step (all dataset statistics reported in this paper are after this step too).}

\mypar{Semantic and Transcription Accuracy.}
In this section we quantify
(i) how well the noun phrases and verbs in the caption correctly represent the objects in the image (Semantic accuracy)
and
(ii) how well the manually transcribed caption matches the voice recording (Transcription accuracy).
We manually check every word in \attcr{\num{100}} randomly selected Localized
Narratives \newcr{on \coco{}} and log each of these two types of errors.
This was done carefully by experts (authors of this paper),
not by the annotators themselves (hence an independent source).

In terms of semantic accuracy, we check every noun and verb in the \attcr{\num{100}} captions and
assess whether that object or action is indeed present in the image.
We allow generality up to a base class name (\eg{} we count either ``dog'' or ``Chihuahua'' as correct for a Chihuahua in the image) and we strictly enforce correctness (\eg{} we count ``skating'' as incorrect when the correct term is ``snowboarding'' or ``bottle'' in the case of a ``jar'').
Under these criteria, semantic accuracy is very high: \attcr{\num{98.0}\%} of the \attcr{\num{1582}} nouns and verbs are accurate.

In terms of transcription accuracy, we listen to the voice recordings
and compare them to the manual transcriptions.
We count every instance of
(i) a missing word in the transcription,
(ii) an extra word in the transcription,
and
(iii) a word with typographical errors.
We normalize these by the total number of words in the
\attcr{\num{100}} captions (\attcr{\num{4059}}).
This results in \attcr{\num{3.3}\%} for type (i), \attcr{\num{2.2}\%} for (ii), and
\attcr{\num{1.1}\%} for (iii), showing transcription accuracy is high.

\begin{figure}[t!]
\resizebox{\linewidth}{!}{%
\mbox{%
\begin{minipage}{0.35\linewidth}
    \mbox{%
\begin{minipage}{\linewidth}
\resizebox{\linewidth}{!}{%
\begin{tikzpicture}
  \begin{axis}[width=1.4\linewidth, height=1.4\linewidth,
  view={0}{90},
  point meta min=0,
  point meta max=10100,
  axis equal=true,
  no markers,
  colorbar,
  colorbar style={ytick=\empty},
  xtick={-1,0,1},
  xticklabels={-1,0,1},
  ytick={-1,0,1},
  yticklabels={-1,0,1}]
    \addplot3 [surf, mesh/rows=49, shader=interp] file {figures/data/2D_hist_trace_location.txt};
    \addplot+[line width=1.5pt,red,dashed] coordinates {(-1,-1) (-1,1) (1,1) (1,-1)(-1,-1)};
    \label{fig:location_hist:box}
  \end{axis}
\end{tikzpicture}}
\end{minipage}
}
    \vspace{-2mm}
    \caption{\small\textbf{Mouse trace segment locations} \newcr{on \coco{}} with respect to
    the closest box of the relevant class (\ref{fig:location_hist:box}).}
    \label{fig:location_hist}
    \vspace{-1mm}
\end{minipage}
\hspace{2mm}
\begin{minipage}{0.65\linewidth}
    \vspace{-4.5mm}
    \centering
\resizebox{\linewidth}{!}{%
\begin{tikzpicture}
  \begin{axis}[width=11cm, height=55mm,
  ylabel=Percentage of captions (\%),ylabel shift={-3pt},xlabel=Number of nouns per caption,xlabel shift={-3pt},
  legend pos = north east,
  minor grid style={white!85!black},
  major grid style={white!60!black},
  xmin=0, xmax=35,
  ymin=0, ymax=55,
  ytick={0,5,...,100},
  xtick={0,5,...,45},
  minor ytick={0,1,...,100},
  minor xtick={0,1,...,100},
  clip marker paths=true,
  enlargelimits=false,grid=both,grid style=densely dotted]
    \addplot+[blue,solid,mark=*, mark size=0.8, line width=1.5pt] coordinates {
    (0,3.187053918450993) (1,24.962900148399406) (2,52.44859020563918) (3,16.405201045862484) (4,2.6535227192424564) (5,0.3427319624054837)
    };
    \addlegendentry{\vg{}}

    \addplot+[black,solid,mark=*, mark size=0.8, line width=1.5pt] coordinates {
    (0,0.030006001200240048) (1,0.9701940388077616) (2,13.432686537307461) (3,37.82756551310262) (4,31.00620124024805) (5,11.88237647529506) (6,3.4906981396279257) (7,0.9401880376075216) (8,0.24004800960192038) (9,0.14002800560112022) (10,0.040008001600320066)
    };
    \addlegendentry{\coco{} Captions}

    \addplot+[olive,solid,mark=*, mark size=0.8, line width=1.5pt, mark options={fill=olive}] coordinates {
    (0,0.048001920076803074) (1,1.5260610424416976) (2,15.14660586423457) (3,28.547141885675426) (4,25.245009800392015) (5,15.16060642425697) (6,7.536301452058082) (7,3.5401416056642265) (8,1.6600664026561063) (9,0.8200328013120526) (10,0.3820152806112245) (11,0.18400736029441178) (12,0.10400416016640666) (13,0.04000160006400256) (14,0.02800112004480179) (15,0.016000640025601025) (16,0.006000240009600384) (17,0.006000240009600384) (18,0.004000160006400256)
    };
    \addlegendentry{\flickr{}}

      \addplot+[red,solid,mark=*, mark size=0.8, line width=1.5pt, mark options={fill=red,solid}] coordinates {(0,0.0) (1,0.08000800080008001) (2,0.9500950095009502) (3,3.1003100310031004) (4,5.320532053205321) (5,6.840684068406841) (6,7.880788078807881) (7,9.24092409240924) (8,9.43094309430943) (9,8.410841084108412) (10,8.17081708170817) (11,7.34073407340734) (12,5.860586058605861) (13,5.0705070507050705) (14,4.27042704270427) (15,3.71037103710371) (16,3.2603260326032606) (17,2.42024202420242) (18,1.7701770177017702) (19,1.5301530153015301) (20,1.3001300130013) (21,0.9900990099009901) (22,0.65006500650065) (23,0.58005800580058) (24,0.41004100410041006) (25,0.22002200220022) (26,0.22002200220022) (27,0.21002100210021002) (28,0.23002300230023) (29,0.06000600060006001) (30,0.09000900090009001) (31,0.09000900090009001) (32,0.030003000300030006) (33,0.07000700070007) (34,0.030003000300030006) (35,0.05000500050005) (36,0.030003000300030006) (37,0.010001000100010001) (38,0.0) (39,0.0) (40,0.010001000100010001) (41,0.030003000300030006) (42,0.020002000200020003) (43,0.010001000100010001) (44,0.0) (45,0.0) (46,0.0)};
      \addlegendentry{Localized Narratives}
      
      \addplot+[teal,solid,mark=*, mark size=0.8, line width=1.5pt, mark options={fill=teal,solid}] coordinates {(0,0.0) (1,0.0) (2,0.015337423312883437) (3,0.03578732106339468) (4,0.2096114519427403) (5,0.6595092024539877) (6,1.8251533742331287) (7,3.9366053169734148) (8,5.117586912065439) (9,5.3323108384458076) (10,4.621676891615542) (11,3.7474437627811863) (12,3.072597137014315) (13,3.0930470347648265) (14,3.374233128834356) (15,4.25357873210634) (16,5.224948875255624) (17,6.600204498977505) (18,7.3517382413087935) (19,7.188139059304703) (20,6.855828220858896) (21,5.74642126789366) (22,4.514314928425358) (23,3.8190184049079754) (24,2.7505112474437627) (25,2.1523517382413084) (26,1.5286298568507157) (27,1.2678936605316973) (28,0.8588957055214724) (29,0.8282208588957056) (30,0.6748466257668712) (31,0.5214723926380368) (32,0.3834355828220859) (33,0.35787321063394684) (34,0.3425357873210634) (35,0.23006134969325154) (36,0.1687116564417178) (37,0.21983640081799594) (38,0.17382413087934562) (39,0.13803680981595093) (40,0.13803680981595093) (41,0.11247443762781187) (42,0.081799591002045) (43,0.09202453987730061) (44,0.07668711656441718) (45,0.030674846625766874) (46,0.06134969325153375) (47,0.0408997955010225) (48,0.030674846625766874) (49,0.015337423312883437) (50,0.03578732106339468) (51,0.02044989775051125) (52,0.005112474437627812) (53,0.0) (54,0.010224948875255624) (55,0.005112474437627812) (56,0.005112474437627812) (57,0.005112474437627812) (58,0.005112474437627812) (59,0.005112474437627812) (60,0.005112474437627812) (61,0.0) (62,0.010224948875255624) (63,0.010224948875255624) (64,0.0) (65,0.0) (66,0.0) (67,0.0) (68,0.0) (69,0.0) (70,0.005112474437627812) (71,0.0) (72,0.0) (73,0.0) (74,0.0) (75,0.0) (76,0.0) (77,0.0) (78,0.0) (79,0.0) (80,0.0) (81,0.0)};
      \addlegendentry{Stanford Visual Paragraphs}
  \end{axis}
\end{tikzpicture}}
    \vspace{-6mm}
    \caption{\small\textbf{Distribution of number of nouns per caption}.
    As in Table~\ref{tab:richness}, these counts are per individual caption. 
    }
    \label{fig:nouns_per_capt}
    \vspace{-4mm}
\end{minipage}}}
\vspace{-4mm}
\end{figure}

\vspace{0.5em}\noindent\textbf{Localization Accuracy.} %
To analyze how well the mouse traces match the location of actual objects
in the image, we extract all instances of any of the 80
\coco{} object classes in our captions (exact string matching\newcr{, 600 classes in the case of \oid{}}).
We recover \attcr{\num{146723}} instances \newcr{on \coco{} and} \attcr{\num{374357}} \newcr{on \oid{} train}.
We then associate each mouse trace segment to the closest ground-truth box of its corresponding class.
Figure~\ref{fig:location_hist} displays the 2D histogram of the positions of all
trace segment points with respect to the closest box (\ref{fig:location_hist:box}),
normalized by box size \newcr{for \coco{}}.
We observe that most of the trace points are within the correct bounding box
\newcr{(the figure for \oid{} is near-identical, see App.~\ref{appendix:localization-oid})}.

We attribute the trace points that fall outside the box to two different effects.
First, circling around the objects is commonly used by annotators
(Fig.~\ref{fig:loc_narr_intro_example} and Fig.~\ref{fig:trace_examples}).
This causes the mouse traces to be close to the box, but not inside it.
Second, some annotators sometimes start moving the mouse before they describe
the object, or vice versa.
We see both cases as a research opportunity to better understand the connection
between vision and language.

\vspace{-2mm}\subsection{Dataset Statistics}
\vspace{-3mm}
\mypar{Richness.}
The mean length of the captions we produced is \captionlength{}~words
(Tab.~\ref{tab:datasets}), substantially longer than all previous datasets,
\newcr{except Stanford Visual Paragraphs~\cite{krause17cvpr} (\eg{}\attcr{\num{3.5}}$\times$ longer than the individual \coco{} Captions~\cite{chen15arxiv})}.
\newcr{Both Localized Narratives and Stanford Visual Paragraphs describe an image with a whole paragraph, as opposed to one sentence~\cite{chen15arxiv,sharma18acl,kazemzadeh14emnlp,mao16cvpr,krishna17ijcv,young14tacl}.
However, Localized Narratives additionally provide dense visual grounding via a mouse trace segment for each word, and has annotations for \attcr{40$\times$} more images than Stanford Visual Paragraphs (Tab.\ref{tab:datasets}).}

We also compare in terms of the average number of nouns, pronouns, adjectives,
verbs, and adpositions (prepositions and postpositions, Tab.~\ref{tab:richness}).
We determined this using the spaCy~\cite{spacy} part-of-speech tagger.
Localized Narratives has a higher occurrence \new{per caption} for each of these categories
compared to \newcr{most} %
previous datasets, which indicates that our annotations provide
richer use of natural language in connection to the images they describe.

\begin{table}[b]
    \vspace{-2mm}
    \centering
    \resizebox{\linewidth}{!}{%
    \begin{tabular}{l@{\hspace{5mm}}c@{\hspace{4mm}}c@{\hspace{4mm}}c@{\hspace{4mm}}c@{\hspace{4mm}}c@{\hspace{4mm}}c}
        \toprule
        Dataset & Words & Nouns & Pronouns & Adjectives & Adpositions & Verbs\\
        \midrule
        \new{\vg{}~\cite{krishna17ijcv}}  & \num{5.1}   & \num{1.9}        & \num{0.0}       & \num{0.6}       &  \num{0.7}      &  \num{0.3}  \\
        \coco{} Captions~\cite{chen15arxiv} & \num{10.5}        & \num{3.6}        & \num{0.2}       & \num{0.8}       &  \num{1.7}      &  \num{0.9}      \\
        \flickr{}~\cite{young14tacl}        & \num{12.4}        & \num{3.9}        & \num{0.3}       & \num{1.1}       &  \num{1.8}      &  \num{1.4}      \\
        \newcr{Localized Narratives}              & \captionlength{}  & \attcr{10.8} & \attcr{3.6} & \attcr{1.6} &  \attcr{4.7} &  \attcr{4.2}\\
        \newcr{Stanford Visual Paragraphs~\cite{krause17cvpr}}          & \attcr{61.9}  & \attcr{17.0} & \attcr{2.7} & \attcr{6.6} &  \attcr{8.0} &  \attcr{4.1}\\
        \bottomrule
    \end{tabular}}\\[2mm]
    \caption{\small\textbf{Richness of individual captions} of Localized Narratives versus previous works.
    \new{Please note that since \coco{} Captions and Flickr30K have replication 5 (and \vg{} also has a high replication), counts \textit{per image} would be higher in these datasets.
    However, many of them would be duplicates.
    We want to highlight the richness of captions as units
    and thus we show word counts averaged over \textit{individual captions}.
}}
    \vspace{-5mm}
    \label{tab:richness}
\end{table}

\mypar{Diversity.}
To illustrate the diversity of our captions, we plot the
distribution of the number of nouns per caption, and compare it to the
distributions obtained over previous datasets (Fig.~\ref{fig:nouns_per_capt}).
\newcr{We observe that the \new{range of} number of nouns is significantly higher in
Localized Narratives than in \coco{} Captions, \flickr{}, \vg{}, and comparable to Stanford Visual Paragraphs.}
This poses an additional challenge for captioning methods: automatically adapting the length of
the descriptions to each image, as a function of the richness of its content.
Beyond nouns, Localized Narratives provide visual grounding for every word (verbs, prepositions, etc.\@).
This is especially interesting for relationship words,
\eg{}``woman holding ballon'' (Fig.~\ref{fig:loc_narr_intro_example}) or
``with a hand under his chin'' (Fig.~\ref{fig:sample_datasets}(d)).
This opens the door to a new venue of research: understanding how humans
naturally ground visual relationships.

\begin{figure}[t]
    \centering
    \setlength{\fboxsep}{0pt}
    \setlength{\fboxrule}{0.7pt}
    \newcommand{\height}{3cm}
    \resizebox{\linewidth}{!}{
    \begin{tabular}{@{}c@{\hspace{1mm}}c@{\hspace{1mm}}c@{\hspace{1mm}}c@{}}
        \fbox{\includegraphics[height=\height]{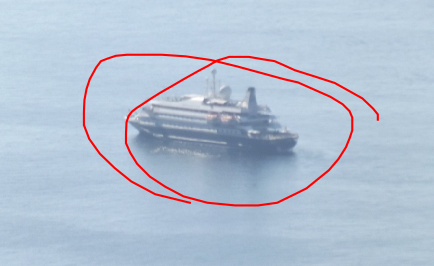}}&
        \fbox{\includegraphics[height=\height]{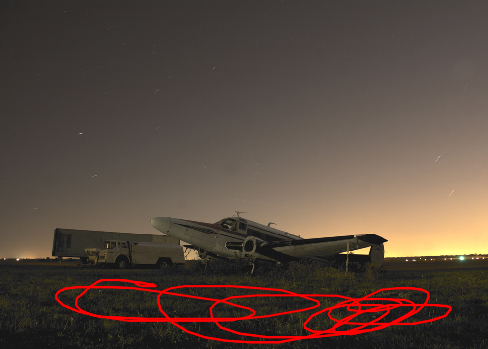}}&
        \fbox{\includegraphics[height=\height]{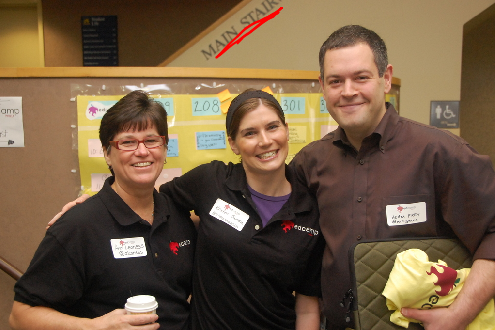}}\\
        Ship & Open land with some grass on it & Main stairs
    \end{tabular}}\\[-2mm]
    \caption{\small\textbf{Examples of mouse trace segments and their corresponding word(s) in the caption} with
    different pointing styles: circling, scribbling, and underlining.}
    \vspace{-3mm}
    \label{fig:trace_examples}
\end{figure}

Diversity in Localized Narratives is present not only in the language modality, but
also in the visual modality, such as the different ways to indicate the spatial location of objects in an image.
In contrast to previous works, where the grounding is in the form of a bounding box,
our instructions lets the annotator hover the mouse over the object in any way they feel natural.
This leads to diverse styles of creating trace segments (Fig.~\ref{fig:trace_examples}):
circling around an object (sometimes without even intersecting it),
scribbling over it, underlining in case of text, etc.
This diversity also presents another challenge: detect and adapt to different trace
styles in order to make full use of them.

\section{Controlled Image Captioning}
\vspace{-2mm}
\label{sec:controlled-image-captioning}
We now showcase how localized narratives can be used for controlled image captioning. 
Controlled captioning was first proposed in~\cite{cornia19cvpr} and enables a user to specify which parts of the image they want to be described, and in which order.
In~\cite{cornia19cvpr} the user input was in the form of user-provided bounding boxes. In this paper we enable controllability through a mouse trace, which provides a more intuitive and efficient user interface.
One especially useful application for controlled captioning is assistive technology for \newcr{people with imperfect vision}~\cite{bigham10uist,wu17cscwsc,zhao17hci}, who could utilize the mouse to express their preferences in terms of how the image description should be presented.

\mypar{Task definition.} Given both an image and a mouse trace, the goal is to produce an image caption which matches the mouse trace, \ie{}it describes the image regions covered by the trace, and in the order of the trace.
This task is illustrated by several qualitative examples of our controlled captioning system in Fig.~\ref{fig:controlled_captioning_2}. In both the image of the skiers and the factory, the caption correctly matches the given mouse trace: it describes the objects which were indicated by the user, in the order which the user wanted.

\begin{figure}[t]
  \vspace{-2mm}
  \centering
  \includegraphics[width=\linewidth]{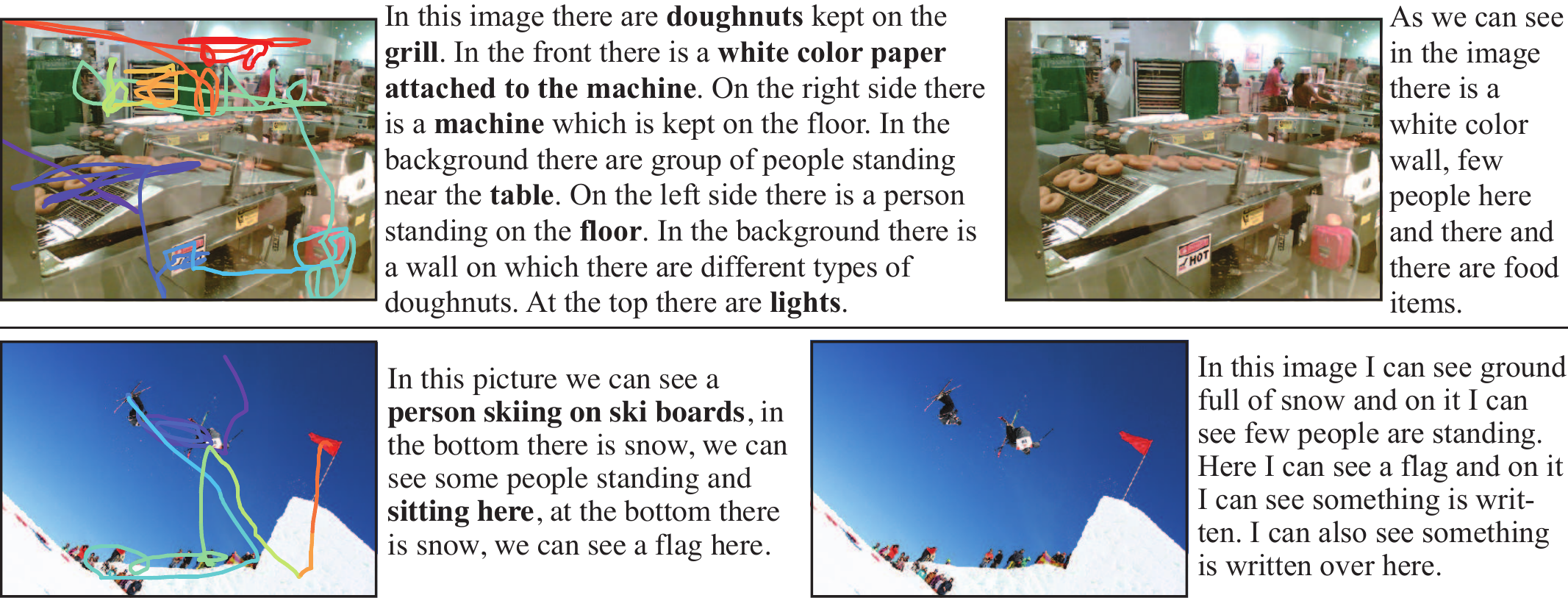}
  \caption{\small\textbf{Qualitative results for controlled image captioning.}
  Gradient \protect\includegraphics[width=8mm]{figures/gradient.pdf} indicates time.
  Captioning controlled by mouse traces (left) and without traces (right).
  The latter misses important objects:
  \eg{}skiers in the sky, doughnuts -- all in bold.}
  \label{fig:controlled_captioning_2}
  \vspace{-7mm}
\end{figure}

\begin{figure}[t!]
  \vspace{-2mm}
  \centering
  \includegraphics[width=0.95\linewidth]{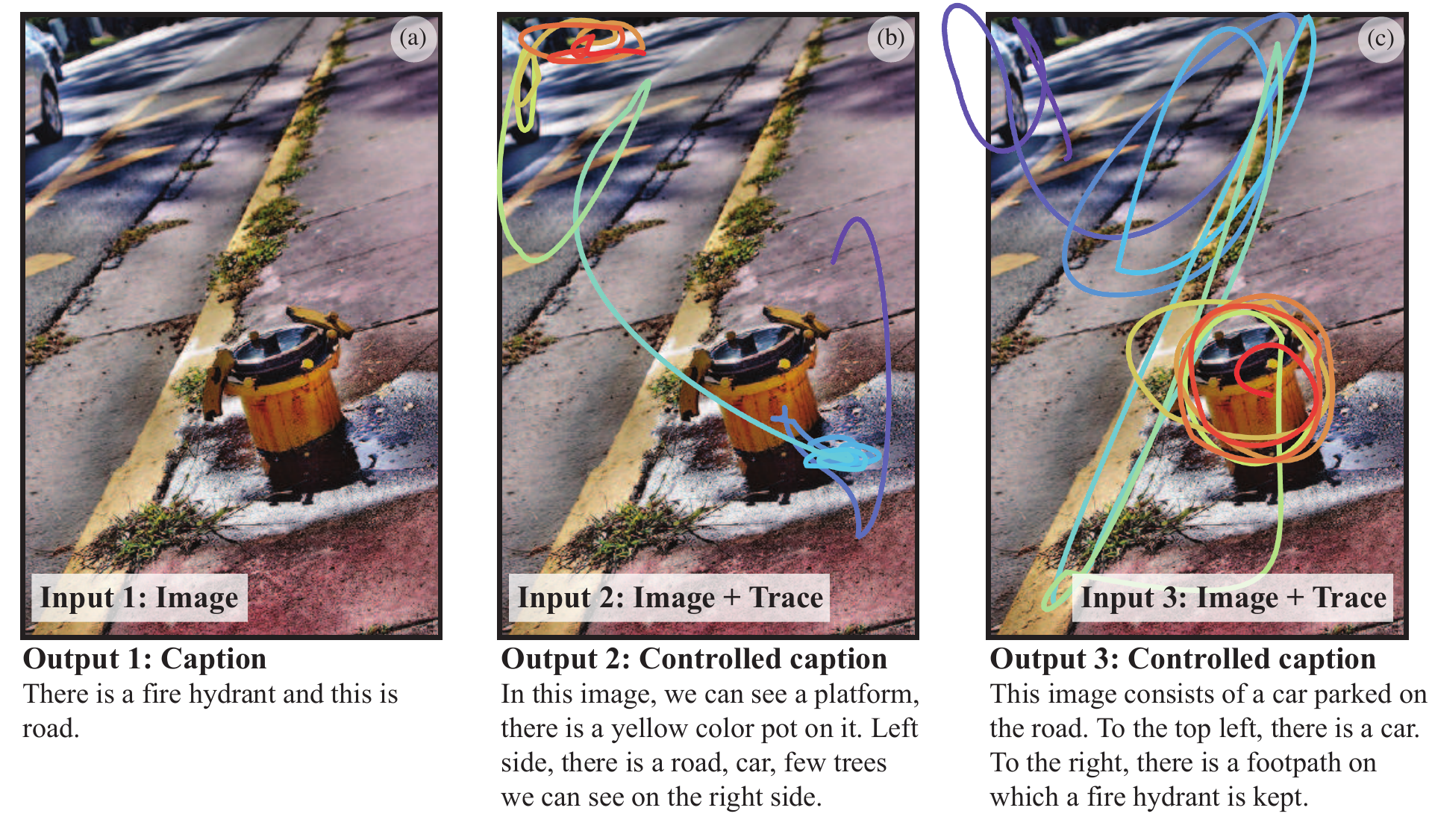}\vspace{-4mm}
  \caption{\small\textbf{Qualitative results for controlled image captioning.} \new{Standard (a) versus controlled captioning (b) and (c).
  (a) misses important objects such as the car or the footpath.
  In (b) and (c) the controlled output captions adapt to the order of the objects
  defined by the trace.
  Gradient \protect\includegraphics[width=8mm]{figures/gradient.pdf} indicates time.} \newcr{More in the App.~\ref{appendix:controlled-captioning-examples}.}}
  \label{fig:controlled_captioning}
  \vspace{-3mm}
\end{figure}

\mypar{Method.}
We start from a state-of-the-art, transformer-based encoder-decoder image
captioning model~\cite{anderson18cvpr,changpinyo2019decoupled}.
This captioning model consumes Faster-RCNN features~\cite{ren15nips} of the top 16 highest
  scored object proposals in the image.
The Faster-RCNN module is pre-trained on \vg{}~\cite{krishna17ijcv}
(excluding its intersection with \coco{}).
The model uses these features to predict an image caption based on an attention
  model, inspired by the Bottom-Up Top-Down approach of~\cite{anderson18cvpr}.
This model is state of the art for standard image captioning,
\ie{}it produces captions given images alone as input (Fig.~\ref{fig:controlled_captioning}(a)).

We modify this model to also input the mouse trace, resulting in
a model that consumes four types of features both at training time and test time:
(i) Faster R-CNN features of the automatically-detected top object proposals,
  representing their semantic information;
(ii) the coordinate and size features of these proposals, representing the
  location of the detected objects.
(iii) the total time duration of the mouse trace, capturing information about
  the expected length of the full image description.
(iv) the position of the mouse trace as it moves over the image, representing
  the visual grounding.
  To create this representation, we first divide the mouse trace evenly into
  pseudo-segments based on the prior median word duration (0.4 sec over the whole
  training set).
  We then represent each pseudo-segment by its encapsulating bounding box,
  resulting in a set of features which take the same form as (ii).
This new model takes an image \textit{plus} a mouse trace
as input and produces the caption that the user is interested in.
\newcr{More technical details in App.~\ref{appendix:controlled-captioning-details}.}

\mypar{Evaluation.} Our first metric is the standard ROUGE-L~\cite{rouge}. This metric determines the longest common subsequence (LCS) of words between the predicted caption and the reference caption, and calculates the F1-score (harmonic mean over precision and recall of words in the LCS).
This means ROUGE-L explicitly measures word order. We also measure the F1 score of ROUGE-1, which we term ROUGE-1-F1. This measures the F1-score w.r.t. co-occurring words. Hence ROUGE-1-F1 is the orderless counterpart of ROUGE-L and enables us to separate the effects of caption completeness (the image parts which the user wanted to be described) and word order (the order in which the user wanted the image to be described).
For completeness we also report other standard captioning metrics: BLEU-1, BLEU-4~\cite{bleu}, \mbox{CIDEr-D}~\cite{cider}, and SPICE~\cite{spice}.
For all measures, a higher number reflects a better agreement between the caption produced by the model and the ground-truth caption written by the annotator.

We observe that in standard image captioning tasks there typically are multiple reference captions to compare to~\cite{chen15arxiv,cider,young14tacl}, since that task is ambiguous: it is unclear what image parts should be described and in which order.
In contrast, our controlled image captioning task takes away both types of ambiguity, resulting in a much better defined task. As such, in this evaluation we compare to a single reference only: given an image plus a human-provided mouse trace, we use its corresponding human-provided caption as reference.

\mypar{Results.}
\new{We perform experiments on the Localized Narratives collected on \coco{} images,
using the standard 2017 training and validation splits. 
To get a feeling of what a trace can add to image captioning, we first discuss the qualitative examples in Fig.~\ref{fig:controlled_captioning_2} and~\ref{fig:controlled_captioning}. First of all, the trace focuses the model attention on specific parts of the image, leading it to mention objects which would otherwise be missed: 
In the top-left image of Fig.~\ref{fig:controlled_captioning_2}, the trace focuses attention on the skiers, which are identified as such (in contrast to the top-right).
Similarly, the top-left and right of Fig.~\ref{fig:controlled_captioning_2}, using the trace results in focusing on specific details which leads to more complete and more fine-grained descriptions (e.g. doughnuts, grill, machine, lights).
Finally in Fig.~\ref{fig:controlled_captioning}a, the standard captioning model misses the car since it is not prominent in the image. In Fig~\ref{fig:controlled_captioning}b and c instead, the augmented model sees both traces going over the car and produces a caption including it. In this same figure, we can also see that different traces lead to different captions.
These results suggests that conditioning on the trace helps with covering the image more completely and highlighting specific objects within it. At the same time, we can see in all examples that the trace order maps nicely to the word order in the caption, which is the order the user wanted.
}

\begin{table}[t!]
  \centering
  \resizebox{\linewidth}{!}{%
    \begin{tabular}{@{\hspace{3mm}}l@{\hspace{3mm}}l@{\hspace{5mm}}c@{\hspace{2mm}}c@{\hspace{2mm}}c@{\hspace{2mm}}c@{\hspace{2mm}}c@{\hspace{2mm}}c@{\hspace{2mm}}}
    \toprule
    Method & features & ROUGE-L & ROUGE-1-F1 & BLEU-1 & BLEU-4 & CIDEr-D & SPICE \\
    \midrule
    \textbf{Standard captioning}~\cite{anderson18cvpr,changpinyo2019decoupled} & \textbf{i} & \textbf{0.317} & \textbf{0.479} & \textbf{0.322} & \textbf{0.081} & \textbf{0.293} & \textbf{0.257} \\
    + proposal locations & i+ii & 0.318 & 0.482 & 0.323 & 0.082 & 0.295 & 0.257 \\
    + mouse trace duration & i+ii+iii & 0.334 & 0.493 & 0.372 & 0.097 & 0.373 & 0.265 \\
    \textbf{Controlled captioning} & \textbf{i+ii+iii+iv} & \textbf{0.483} & \textbf{0.607} & \textbf{0.522} & \textbf{0.246} & \textbf{1.065} & \textbf{0.365} \\
    \bottomrule
  \end{tabular}}\\[2mm]
  \caption{\small\textbf{Controlled image captioning results on the \coco{} validation set},
  \new{versus standard (non-controlled) captioning, and two ablations.}}
  \label{tab:tbTraceFeat}
  \vspace{-7mm}
\end{table}

Table~\ref{tab:tbTraceFeat} shows quantitative results.
Compared to standard captioning~\cite{anderson18cvpr,changpinyo2019decoupled}, all metrics improve significantly when doing controlled captioning using the mouse trace. BLEU-4 and CIDEr-D are particularly affected and improve by more than \att{\num{3}}$\times$.
ROUGE-1-F1 increased from 0.479 for standard captioning to 0.607 for controlled captioning using the full mouse trace. Since ROUGE-1-F1 ignores word order, this increase is due to the \emph{completeness} of the caption only:
it indicates that using the mouse trace enables the system to better describe those parts of the image which were indicated by the user.

Switching from ROUGE-1-F1 to ROUGE-L imposes a word order. The standard captioning model yields a ROUGE-L of 0.317, a drop of 34\% compared to ROUGE-1-F1. Since standard captioning does not input any particular order within the image (but does use a linguistically plausible ordering), this drop can be seen as a baseline for not having information on the order in which the image should be described.
When using the mouse trace, the controlled captioning model yields a ROUGE-L of 0.483, which is a much smaller drop of 20\%.
This demonstrates quantitatively that our controlled captioning model successfully exploits the input trace to determine the order in which the user wanted the image to be described.
Overall, the controlled captioning model outperforms the standard captioning model by 0.166 ROUGE-L on this task (0.483 vs 0.317).

\mypar{Ablations.}
\new{We perform two ablations to verify whether most improvements indeed come from the mouse trace itself, as opposed to the other features we added.
Starting from standard captioning, we add the locations of the object proposals from which the model extracts visual features (Tab.~\ref{tab:tbTraceFeat}, ``+ proposal locations'', feature (ii)). This has negligible effects on performance, suggesting that this model does not benefit from knowing where in the image its appearance features (i) came from.
Next, we add the trace time duration (Tab.~\ref{tab:tbTraceFeat}, ``+ mouse trace duration''). This gives an indication of how long the caption the user wants should be. This brings minor improvements only.
Hence, most improvements come when using the full mouse trace, demonstrating that most information comes from the location and order of the mouse trace (Tab.~\ref{tab:tbTraceFeat}, controlled captioning).

\mypar{Summary.}
To summarize, we demonstrated that using the mouse trace leads to large improvements when compared to a standard captioning model, for the task of controlled captioning. Importantly, we do not claim the resulting captions are better in absolute terms. Instead, they are better fitting what the user wanted, in terms of which parts of the image are described and in which order.
}

\section{Conclusions}
\label{sec:conclusions}
This paper introduces Localized Narratives, an efficient way to collect image
captions in which every single word is visually grounded by a mouse trace.
We annotated \numimagesshort{} images with Localized Narratives.
Our analysis shows that our data is rich and provides accurate grounding.
We demonstrate the utility of our data through controlled image captioning using the mouse trace.

\bibliographystyle{splncs04}
\bibliography{shortstrings,loco}

\section*{Appendices}
\appendix
The appendices are organized as follows:
\begin{description}
\item[Appendix~\ref{appendix:image-generation-using-traces}] presents a demonstration of application of Localized Narratives for image generation.
The user describes the image they want by means of a Localized Narrative and the method generates an image that matches the description.
\item[Appendix~\ref{appendix:controlled-captioning-examples}] provides additional qualitative examples for the controlled image captioning application.
\item[Appendix~\ref{appendix:localization-oid}] \newcr{provides an additional quantitative plot of the localization accuracy of the mouse trace in Localized Narratives for \oid{}. It was suppressed from the  main paper due to space limitations}.
\item[Appendix~\ref{appendix:controlled-captioning-details}] provides additional technical details on the framework that we use for controlled image captioning.
\end{description}

\section{Image Generation}
\label{appendix:image-generation-using-traces}
Generating an image conditioned on a semantic segmentation map is a well-studied application~\cite{chen17iccv,isola17cvpr,park19cvpr,wang18cvpr}.
However, while such segmentation maps give control over the image to be synthesized, they do not provide a natural interface for the user.
In this section, we show how we can use labelled mouse traces to generate images.
This opens up a new and intuitive way for the user to provide guidance to the image generation process.

We start from SPADE~\cite{park19cvpr}, which is an existing, state-of-the-art framework for generating images conditioned on a pixel-wise segmentation map. We use their publicly available model that is pre-trained on COCO-stuff~\cite{caesar18cvpr,lin14eccv}, which features 182 semantic classes, including object and background classes (stuff).
At test time, the model takes as input a segmentation map where pixels are labeled with these classes, and generates an image.
In this section we exploit Localized Narratives as a natural interface for producing these segmentation maps efficiently, as the user can specify both the location and class label of the desired image elements at the same time, and can intuitively specify elements in their order of importance.

\begin{figure*}[thpb]
\centering
  \setlength{\fboxsep}{0pt}
  \setlength{\fboxrule}{2pt}
  \resizebox{0.99\linewidth}{!}{%
  \fbox{\includegraphics[trim={9pt 9pt 9pt 9pt}, clip]{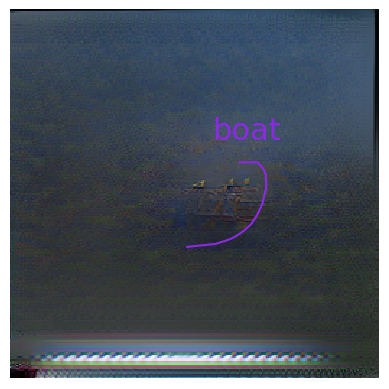}}\hspace{3mm}
  \fbox{\includegraphics[trim={9pt 9pt 9pt 9pt}, clip]{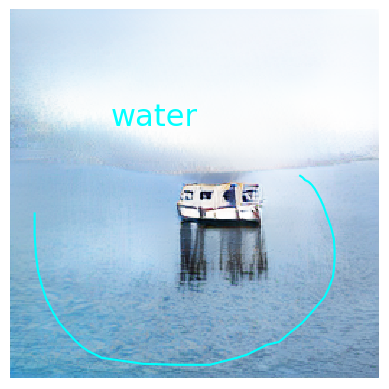}}\hspace{3mm}
  \fbox{\includegraphics[trim={9pt 9pt 9pt 9pt}, clip]{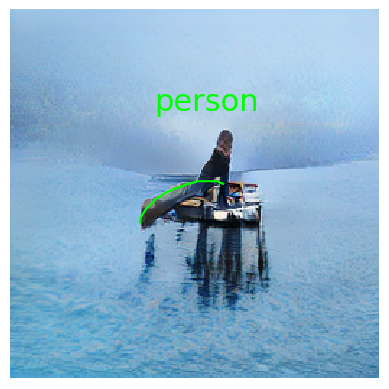}}\hspace{3mm}
  \fbox{\includegraphics[trim={9pt 9pt 9pt 9pt}, clip]{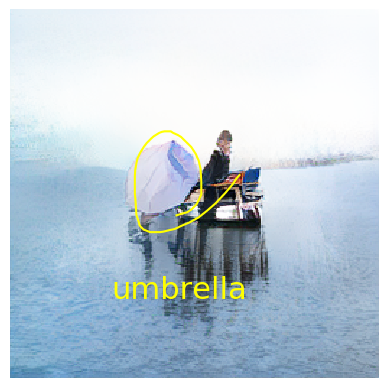}}\hspace{3mm}
  \fbox{\includegraphics[trim={9pt 9pt 9pt 9pt}, clip]{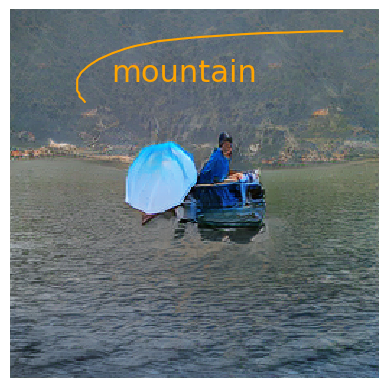}}}\\[1mm]
  \resizebox{.99\linewidth}{!}{%
  \fbox{\includegraphics[trim={9pt 9pt 9pt 9pt}, clip]{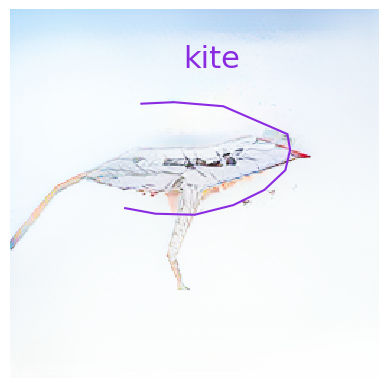}}\hspace{3mm}
  \fbox{\includegraphics[trim={9pt 9pt 9pt 9pt}, clip]{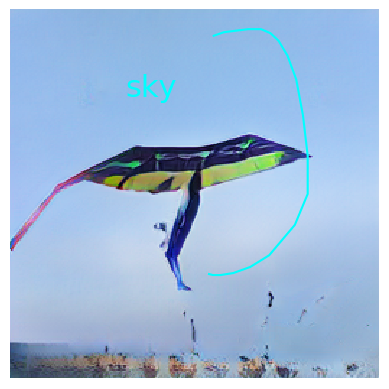}}\hspace{3mm}
  \fbox{\includegraphics[trim={9pt 9pt 9pt 9pt}, clip]{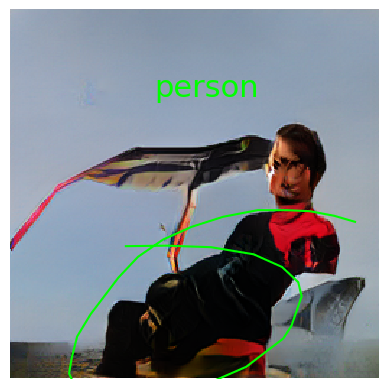}}\hspace{3mm}
  \fbox{\includegraphics[trim={9pt 9pt 9pt 9pt}, clip]{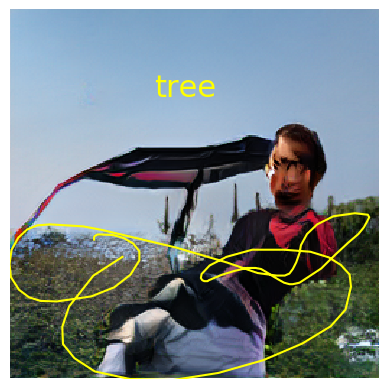}}\hspace{3mm}
  \fbox{\includegraphics[trim={9pt 9pt 9pt 9pt}, clip]{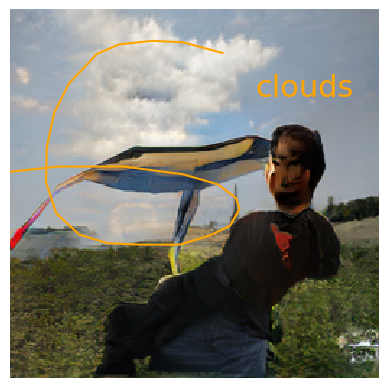}}}\\[1mm]
  \resizebox{.99\linewidth}{!}{%
  \fbox{\includegraphics[trim={9pt 9pt 9pt 9pt}, clip]{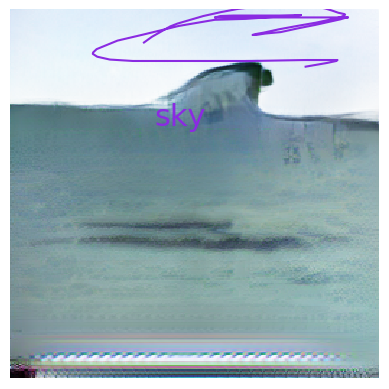}}\hspace{3mm}
  \fbox{\includegraphics[trim={9pt 9pt 9pt 9pt}, clip]{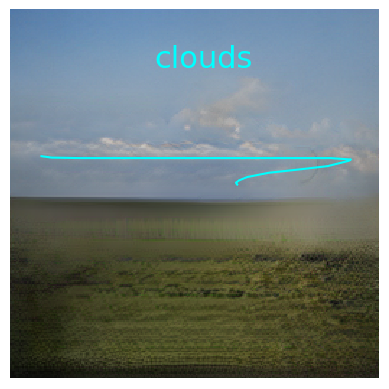}}\hspace{3mm}
  \fbox{\includegraphics[trim={9pt 9pt 9pt 9pt}, clip]{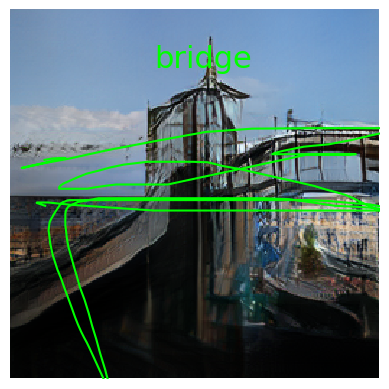}}\hspace{3mm}
  \fbox{\includegraphics[trim={9pt 9pt 9pt 9pt}, clip]{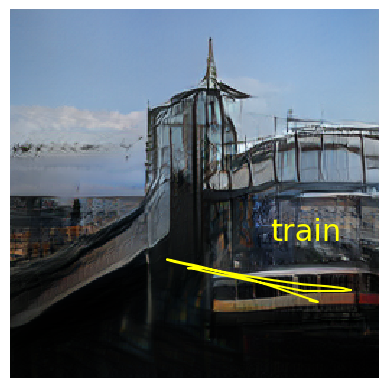}}\hspace{3mm}
  \fbox{\includegraphics[trim={9pt 9pt 9pt 9pt}, clip]{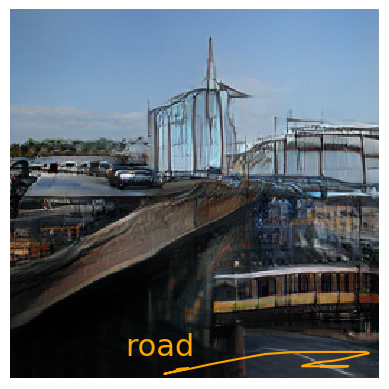}}}\\[1mm]
  \resizebox{.99\linewidth}{!}{%
  \fbox{\includegraphics[trim={9pt 9pt 9pt 9pt}, clip]{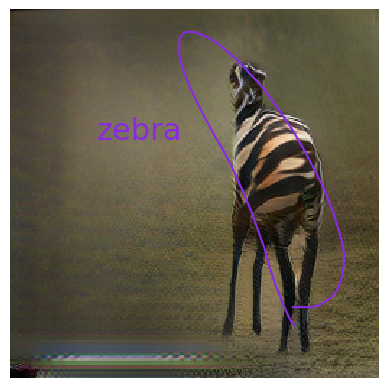}}\hspace{3mm}
  \fbox{\includegraphics[trim={9pt 9pt 9pt 9pt}, clip]{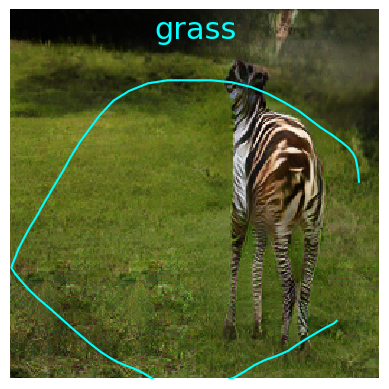}}\hspace{3mm}
  \fbox{\includegraphics[trim={9pt 9pt 9pt 9pt}, clip]{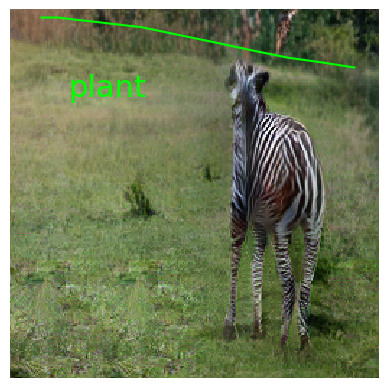}}\hspace{3mm}
  \fbox{\includegraphics[trim={9pt 9pt 9pt 9pt}, clip]{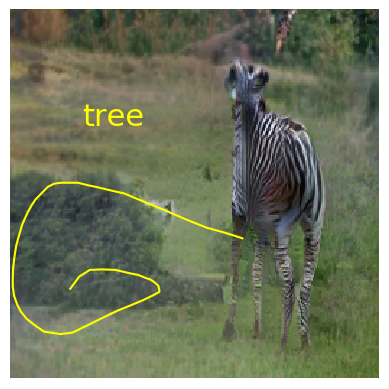}}\hspace{3mm}
  \hphantom{\fbox{\includegraphics[trim={9pt 9pt 9pt 9pt}, clip]{figures/gan_narratives/159977_3_tree.png}}}}\\[1mm]
\resizebox{.99\linewidth}{!}{%
  \fbox{\includegraphics[trim={9pt 9pt 9pt 9pt}, clip]{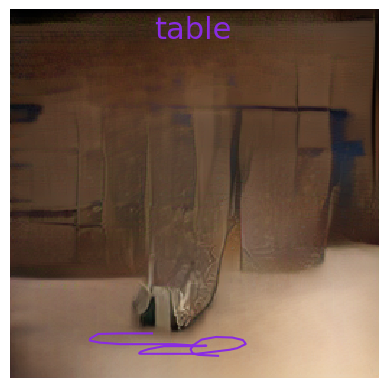}}\hspace{3mm}
  \fbox{\includegraphics[trim={9pt 9pt 9pt 9pt}, clip]{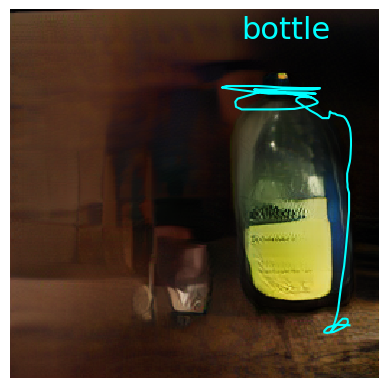}}\hspace{3mm}
  \fbox{\includegraphics[trim={9pt 9pt 9pt 9pt}, clip]{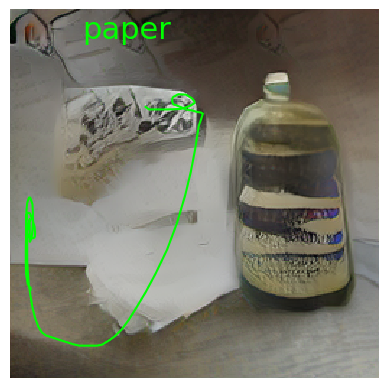}}\hspace{3mm}
  \fbox{\includegraphics[trim={9pt 9pt 9pt 9pt}, clip]{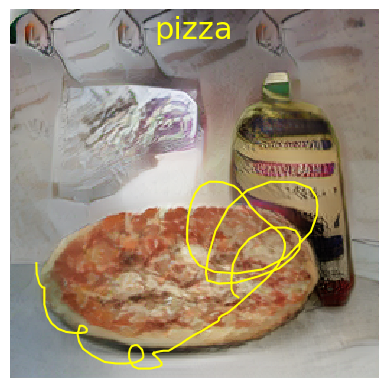}}\hspace{3mm}
  \hphantom{\fbox{\includegraphics[trim={9pt 9pt 9pt 9pt}, clip]{figures/gan_narratives/116208_3_pizza.png}}}}\\[1mm]
\resizebox{.99\linewidth}{!}{%
  \fbox{\includegraphics[trim={9pt 9pt 9pt 9pt}, clip]{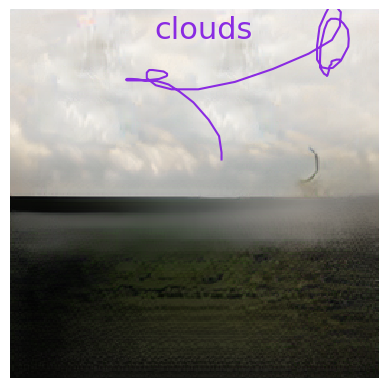}}\hspace{3mm}
  \fbox{\includegraphics[trim={9pt 9pt 9pt 9pt}, clip]{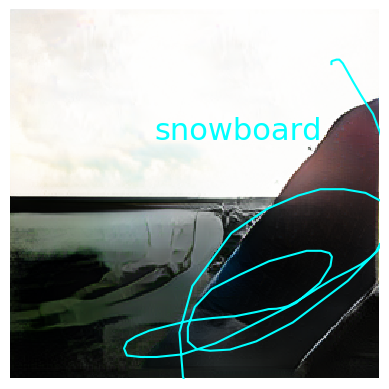}}\hspace{3mm}
  \fbox{\includegraphics[trim={9pt 9pt 9pt 9pt}, clip]{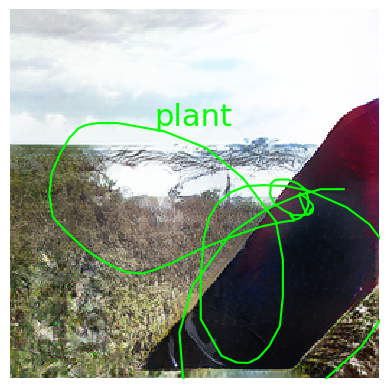}}\hspace{3mm}
  \fbox{\includegraphics[trim={9pt 9pt 9pt 9pt}, clip]{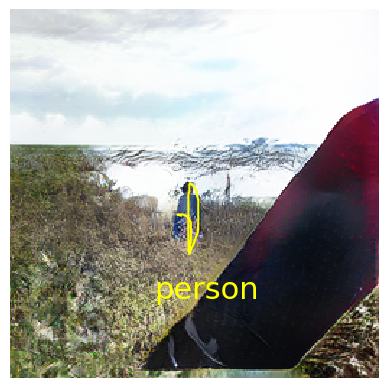}}\hspace{3mm}
  \hphantom{\fbox{\includegraphics[trim={9pt 9pt 9pt 9pt}, clip]{figures/gan_narratives/2532_3_person.png}}}}\\[1mm]
\resizebox{.99\linewidth}{!}{%
  \fbox{\includegraphics[trim={9pt 9pt 9pt 9pt}, clip]{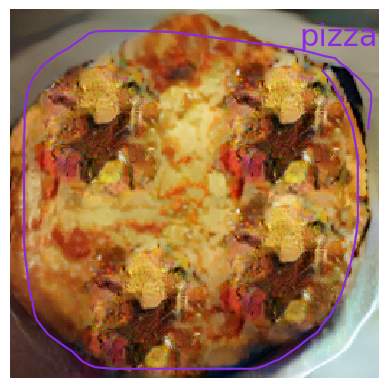}}\hspace{3mm}
\fbox{\includegraphics[trim={9pt 9pt 9pt 9pt}, clip]{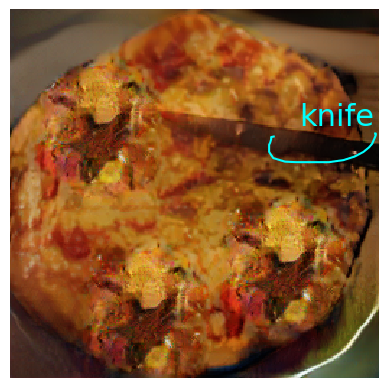}}\hspace{3mm}
  \fbox{\includegraphics[trim={9pt 9pt 9pt 9pt}, clip]{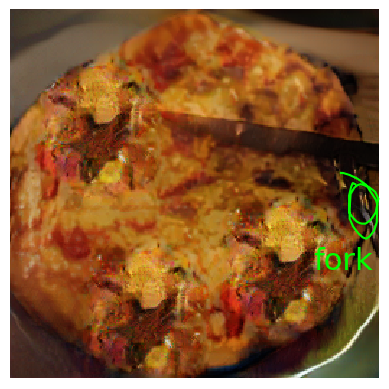}}\hspace{3mm}
  \hphantom{\fbox{\includegraphics[trim={9pt 9pt 9pt 9pt}, clip]{figures/gan_narratives/8629_2_fork.png}}}\hspace{3mm}
  \hphantom{\fbox{\includegraphics[trim={9pt 9pt 9pt 9pt}, clip]{figures/gan_narratives/8629_2_fork.png}}}}
  \caption{\textbf{Seven examples (one per row) of image generation using mouse traces.} New image elements are iteratively added (from left to right) using a noun and its associated trace segment.}
  \label{fig:image_generation}
\end{figure*}

\mypar{Localized Narrative to Semantic Segmentation Map.}
For this application we need to convert the labelled traces into an appropriate segmentation map. We found that scene elements should have a realistic shape for SPADE to produce a pleasing image. Furthermore, SPADE deals poorly with maps which consist mostly of unlabelled pixels. To overcome this, we first collect masks for 1000 instances of each class from the COCO-stuff training set (both object and background classes). Given a trace segment with a class label, we first create its convex hull. We then compare it to all training instances of the same class and select the mask with the highest spatial overlap. This mask has a natural shape since it comes from a real instance.

Equipped with these retrieved masks, we construct a semantic segmentation map.
We start from an empty map where all pixels are unlabelled, and iteratively add masks in the same order as the trace segments.
An object mask is pasted on top of the current map, overwriting any previously labelled pixels.
A background mask only overwrites pixels labeled as another background class.
This approach results in using masks that cover more surface compared to the input trace segments,
which helps reducing the surface of unlabelled pixels.

\mypar{Results.}
Figure~\ref{fig:image_generation} shows seven examples created based on Localized Narratives.
In both examples, the images get increasingly complex as the Localized Narrative continues, while also preserving previous details.
In the first example, the closed boat becomes open once the user indicated that a person should be visible on the boat. Moreover, the addition of the mountain alters the appearance of the water. In the second example, adding the clouds effectively changes the weather conditions and therefore the illumination. The other examples follow similar patterns.

\mypar{Conclusions.}
We demonstrated that Localized Narratives can be used for image generation. Since we kept the pre-trained SPADE~\cite{park19cvpr} model unmodified and only used traces to create segmentation maps, we do not believe our framework generates better images. Instead, we demonstrated that we can generate images incrementally with an intuitive interface. More importantly, while we now only generated \emph{nouns}, Localized Narratives opens up the possibility to also consider adjectives such as \emph{red} or \emph{old} and verbs such as \emph{holding} and \emph{riding}. We feel this presents exciting and challenging new research opportunities.

\section{Additional Qualitative Examples for Controlled Image Captioning}
\label{appendix:controlled-captioning-examples}
\begin{figure}[hptb]
\begin{tabular}{p{0.48\textwidth}@{\hspace{0.04\textwidth}}p{0.48\textwidth}}
  \toprule
  \centering Traditional Captioning & \centering\arraybackslash Controlled Captioning \\
  \midrule
  \twocaptions{4cm}{}{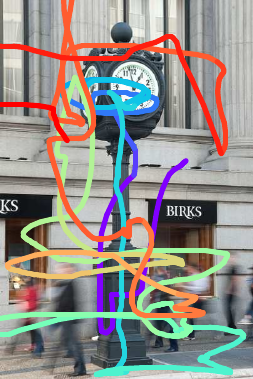}{%
    This is a black and white picture.
    Here we can see clocks on the pole.
    In the background there is a building and this is sky.%
  }{%
    In the center of the image there is a black pole to which clocks are placed.
    At the bottom of the image, we can see a group of people walking on the road.
    In the background, there is a building.}
  \midrule
  \twocaptions{4cm}{}{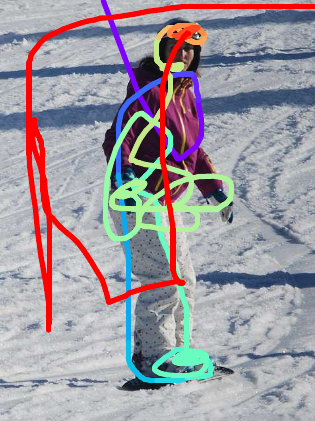}{%
    In this image, there is snow on the ground which is in white color, in the middle
    there is a person standing on the ski board and wearing a red color jacket,
    in the background there are some green color plants.%
  }{%
    Here in this picture we can see a person skiing on snow with ski board on her
    legs and she is also wearing gloves, goggles and a helmet on her and we can
    see the ground is covered with snow over there.}
  \bottomrule
\end{tabular}
\caption{\small\textbf{Controlled Captioning Qualitative Examples 1}: Traditional captioning where the input is only the image (left)
  versus our captioning controlled by mouse traces where the mouse traces are also an input to the model (right). Gradient \protect\includegraphics[width=8mm]{figures/gradient.pdf} indicates time.}
\label{fig:controlled_captioning_1}
\end{figure}

  Figure~\ref{fig:controlled_captioning_1} and
  Figure~\ref{fig:controlled_captioning_2} show additional qualitative examples of
  controlled versus classical image captioning on our data.

\begin{figure}[hptb]
  \begin{tabular}{p{0.48\textwidth}@{\hspace{0.04\textwidth}}p{0.48\textwidth}}
    \toprule
    \centering Traditional Captioning & \centering\arraybackslash Controlled Captioning \\
    \midrule
    \twocaptions{\linewidth}{}{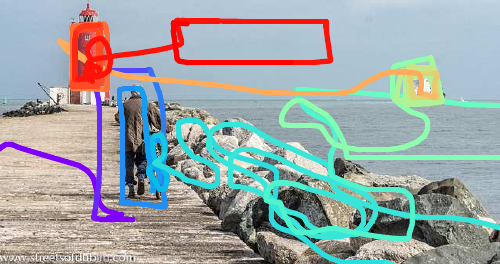}{%
      In this picture we can see a man wore jacket holding bicycle with his hand
      and beside to him we can see rocks, water, ship and in the background we can see sky.
    }{%
      In this image we can see a man standing on the left side.
      He is holding a bicycle in his hand.
      Here we can see stones on the right side.
      Here we can see a ship on the top right side.
      Here we can see a tower on the left side.
      This is a sky.%
    }
    \midrule
    \twocaptions{4cm}{}{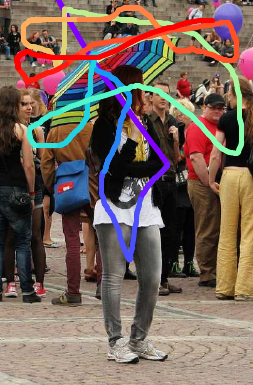}{%
      In this picture we can see a group of persons standing on the ground and
      in the background we can see a building, trees, sky.
    }{%
      A person is standing wearing a black dress and holding a umbrella.
      Behind her there are other people standing.
      At the left and right there are kites.
      There are trees at the back.%
    }
    \bottomrule
  \end{tabular}
  \caption{\small\textbf{Controlled Captioning Qualitative Examples 2}: Traditional captioning where the input is only the image (left)
    versus our captioning controlled by mouse traces where the mouse traces are also an input to the model (right). Gradient \protect\includegraphics[width=8mm]{figures/gradient.pdf} indicates time.}
  \label{fig:controlled_captioning_2}
\end{figure}

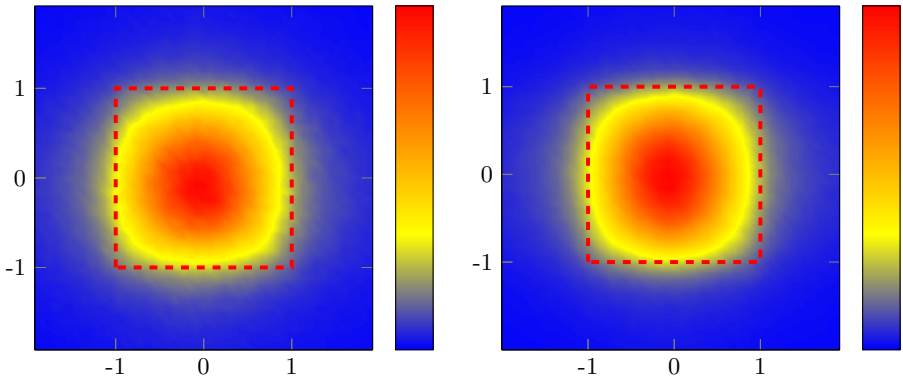
\begin{figure}[hptb]
\begin{tikzpicture}
  \begin{axis}[width=0.5\linewidth, height=0.5\linewidth,
  view={0}{90},
  point meta min=0,
  point meta max=10100,
  axis equal=true,
  no markers,
  colorbar,
  colorbar style={ytick=\empty},
  xtick={-1,0,1},
  xticklabels={-1,0,1},
  ytick={-1,0,1},
  yticklabels={-1,0,1}]
    \addplot3 [surf, mesh/rows=49, shader=interp] file {figures/data/2D_hist_trace_location.txt};
    \addplot+[line width=1.5pt,red,dashed] coordinates {(-1,-1) (-1,1) (1,1) (1,-1)(-1,-1)};
    \label{fig:location_hist:box}
  \end{axis}
\end{tikzpicture}
\hspace{2mm}
\begin{tikzpicture}
  \begin{axis}[width=0.5\linewidth, height=0.5\linewidth,
  view={0}{90},
  point meta min=0,
  point meta max=23832,
  axis equal=true,
  no markers,
  colorbar,
  colorbar style={ytick=\empty},
  xtick={-1,0,1},
  xticklabels={-1,0,1},
  ytick={-1,0,1},
  yticklabels={-1,0,1}]
    \addplot3 [surf, mesh/rows=50, shader=interp] file {figures/data/open_images_train_2D_hist_trace_location.txt};
    \addplot+[line width=1.5pt,red,dashed] coordinates {(-1,-1) (-1,1) (1,1) (1,-1)(-1,-1)};
    \label{fig:location_hist:box}
  \end{axis}
\end{tikzpicture}
  \caption{\small\textbf{Histograms of mouse trace segment locations} on \coco{} (left) and \oid{} (right) with respect to the closest box of the relevant class (\ref{fig:location_hist:box}).}
  \label{fig:location_histograms}
\end{figure}

\newcr{\section{Localization accuracy on \oid{}}\label{appendix:localization-oid}}
Figure~\ref{fig:location_histograms} shows the histograms of mouse trace segment locations on \coco{} (left) and \oid{} (right) with respect to the closest box of the relevant class (The main paper also shows the histogram for \coco{}).

\section{Controlled Image Captioning Details}
\label{appendix:controlled-captioning-details}
\subsection{Method and Training Details}\label{subsec:method-and-training-details}

Our transformer-based encoder-decoder image captioning model follows the
architecture in~\cite{changpinyo2019decoupled} with a few minor differences.
First, we set the number of Transformers' layers for both the encoder and the
decoder to \num{2} instead of \num{6}.
Second, our projection layers also consist of layer normalization~\cite{ba2016layer}
(Sec.~\ref{subsec:controlled-captioning:rep}).
Third, we set the maximum number of iterations to \num{150}k, much smaller than the \num{2}M used in that work.
Finally, we allow the maximum number of target captions to be as long as \num{225} to account for the longer nature of the narration.

Besides above, our input features are standard regional Faster
R-CNN~\cite{ren15nips} features: no ultra-finegrained, global, or entity
features are involved.
We will describe how we represent these and additional features in Section~\ref{subsec:controlled-captioning:rep}.

\subsection{Representations of visual and trace features}
\label{subsec:controlled-captioning:rep}

Recall from the main text that our model consumes up to four types of features:
(i) Faster R-CNN features of the automatically-detected top object proposals, representing their semantic information;
(ii) the coordinate and size features of these proposals, representing the location of the detected objects.
(iii) the total time duration of the mouse trace, capturing information about the expected length of the full image description.
(iv) the position of the mouse trace as it moves over the image, representing the visual grounding.
To create this representation, we first divide the mouse trace evenly into pseudo-segments based on the prior median word duration (\num{0.4} sec over the whole training set).
We then represent each pseudo-segment by its encapsulating bounding box, resulting in a set of features which take the same form as (ii).

\paragraph{Visual Features.}
Faster R-CNN features (i) are represented by a sequence of $R$=16 \num{2048}D vectors: $f_1, f_2, \ldots, f_R$ (output by the Faster R-CNN), which are later projected onto a \num{512}D vector and followed by layer normalization.

We represent the location of detected objects (ii) with a sequence of \num{5}D vectors: $p_1, p_2, \ldots, p_R$.
Each vector contains numbers between \num{0} and \num{1} corresponding to the
top-left $x$ and $y$ coordinates, the bottom-right $x$ and $y$ coordinates,
and the area with respect to the whole image.
We project it to a \num{512}D space as for (i) above.

To construct a representation of (i + ii), we add the projected and normalized
vectors from each source and apply another layer normalization to the
resulting vector, leading to a sequence of $R$ \num{512}D vectors.

For the visual features above, we do not use ``time'' positional encoding such
that the model is permutation-invariant to the sequence vectors.

\paragraph{Trace Features.}
As mentioned in the main text, the mouse trace coordinates are uniformly
divided into a 0.4-second pseudo-segments of trace coordinates and then
converted into a series of corresponding bounding boxes.
Thus, we now have a sequence of \num{5}D vectors: $q_1, q_2, \ldots, q_T$,
where $q_j$ has the same form as (ii).

Each box corresponds to the smallest region that covers the trace,
which might potentially not cover the whole object.
To mitigate this, we extend the box in each direction by the offset $\delta$.
To represent (iii), we set $\delta=1.0$ such as $q_j$'s become all
[0, 0, 1, 1, 1] (the region of the whole image).
In other words, all the trace position information is dismissed,
leaving only the total time duration of the mouse trace.
On the other hand, setting $\delta=0.1$ gives the (iii + iv) signal.
After the transformation, we have a sequence of \emph{transformed} \num{5}D vectors:
$q'_1, q'_2, \ldots, q'_T$, which are later projected onto a \num{512}D vector and
followed by layer normalization.

Different from the visual features, each trace comes with the notion of ``time'' --- the order of the regions that are derived from traces matters.
Thus, we construct such a time representation $\mathrm{sinusoids}(1), \mathrm{sinusoids}(2), \ldots, \mathrm{sinusoids}(T)$,
where $\mathrm{sinusoids}(j)$ is a \num{512}D vector of $j$ based on sine and
cosine functions of different frequencies~\cite{vaswani2017attention}.
Similarly to (i + ii), when combining the the trace features with sinusoids,
we add the vectors from each source and apply another layer normalization
to the resulting vector.
In the end, we have a sequence of $T$ \num{512}D vectors.

\paragraph{Combining Visual and Trace Features.}
To construct (i + ii + iii) or (i + ii + iii + iv), we simply concatenate
the ``visual feature'' sequence and the ``trace feature''
sequence and use the result as the input to the model.

\end{document}